\documentclass[10pt,twocolumn,letterpaper]{article}

\usepackage{cvpr}      % To produce the REVIEW version
\usepackage{url}

\usepackage{graphicx}
\usepackage{multirow}
\usepackage{subcaption}
\usepackage{dblfloatfix}
\usepackage{booktabs}
\usepackage{colortbl}
\usepackage{enumitem}

\usepackage{xcolor}
\definecolor{mygray}{gray}{.9}
\definecolor{mygreen}{RGB}{34,139,34}

\newcommand{\best}[1]{{\textbf{\@#1}}}

\newcommand{\model}{Mod-Squad\xspace}

\usepackage{times}
\usepackage{epsfig}
\usepackage{graphicx}
\usepackage{amsmath}
\usepackage{amssymb}
\usepackage{cite}

\usepackage[pagebackref=true,breaklinks=true,colorlinks,citecolor=mygreen,bookmarks=false]{hyperref}
\usepackage{graphicx}
\usepackage{amsmath}
\usepackage{amssymb}
\usepackage{booktabs}

\usepackage[capitalize]{cleveref}
\crefname{section}{Sec.}{Secs.}
\Crefname{section}{Section}{Sections}
\Crefname{table}{Table}{Tables}
\crefname{table}{Tab.}{Tabs.}

\def\cvprPaperID{6727} % *** Enter the CVPR Paper ID here
\def\confName{CVPR}
\def\confYear{2023}

\begin{document}

%%%%%%%%% TITLE - PLEASE UPDATE
%\title{Train Once, Get All: \\ Learning to Compose All Tasks Within One Model}

% Cooperation and Specialization: \\ Designing Mixture of Experts to be Modularized Multi-Task Learners\\
\title{
Mod-Squad: Designing Mixture of Experts As Modular Multi-Task Learners
}
% Mod-Squad: Mixtures of Modularized Learners
% Mod-Squad: Modular mixture of experts
% Mod-Squad: Mixture of Squads Are Strong Modularized Multi-Task Learner
% \author{First Author\\
% Institution1\\
% Institution1 address\\
% {\tt\small firstauthor@i1.org}
% % For a paper whose authors are all at the same institution,
% % omit the following lines up until the closing ``}''.
% % Additional authors and addresses can be added with ``\and'',
% % just like the second author.
% % To save space, use either the email address or home page, not both
% \and
% Second Author\\
% Institution2\\
% First line of institution2 address\\
% {\tt\small secondauthor@i2.org}
% }

\author{Zitian Chen\textsuperscript{1}, Yikang Shen\textsuperscript{2}, Mingyu Ding\textsuperscript{3}, Zhenfang Chen\textsuperscript{2},\\ Hengshuang Zhao\textsuperscript{3}, Erik Learned-Miller\textsuperscript{1}, Chuang Gan\textsuperscript{1,2} \\
        {
            \small \textsuperscript{1} University of Massachusetts Amherst, \textsuperscript{2} MIT-IBM Watson AI Lab, \textsuperscript{3} The University of Hong Kong 
        }
        }
        
\maketitle

%%%%%%%%% ABSTRACT
\begin{abstract}
   \indent Optimization in multi-task learning (MTL) is more challenging than single-task learning (STL), as the gradient from different tasks can be contradictory.  When tasks are related, it can be beneficial to share some parameters among them (cooperation). However, some tasks require additional parameters with expertise in a specific type of data or discrimination (specialization). 
   To address the MTL challenge, we propose \textbf{\model}, a new model that is {\bf Mod}ularized into groups of experts (a `{\bf Squad}'). This structure allows us to formalize cooperation and specialization as the process of matching experts and tasks. We optimize this matching process during the training of a single model. Specifically, we incorporate mixture of experts (MoE) layers into a transformer model, with a new loss that incorporates  the
    mutual dependence between tasks and experts. As a result, only a small set of experts are activated for each task. This prevents the sharing of the entire backbone model between all tasks, which strengthens the model, especially when the training set size and the number of tasks scale up.  
    % More interestingly, the small sub-network from this large model can work independently as a standalone model for individual task with no performance drop. 
    More interestingly, for each task, we can extract the small set of experts as a standalone model that maintains the same performance as the large model.
    Extensive experiments on the Taskonomy dataset with 13 vision tasks and the PASCAL-Context dataset with 5 vision tasks show the superiority of our approach.
%   In this process, MoE layers automatically models the relation between tasks. 
   
\end{abstract}
% Why we need the loss? By maximize mutual information between tasks and experts, the experts specialize for the task while using the minimum capacity. 
% Mutual dependence between model and task

%%%%%%%%% BODY TEXT
\section{Introduction}
\label{sec:intro}

\begin{figure}[t]
    % \hspace{-0.5in}
  \centering
  \includegraphics[width=8.5cm]{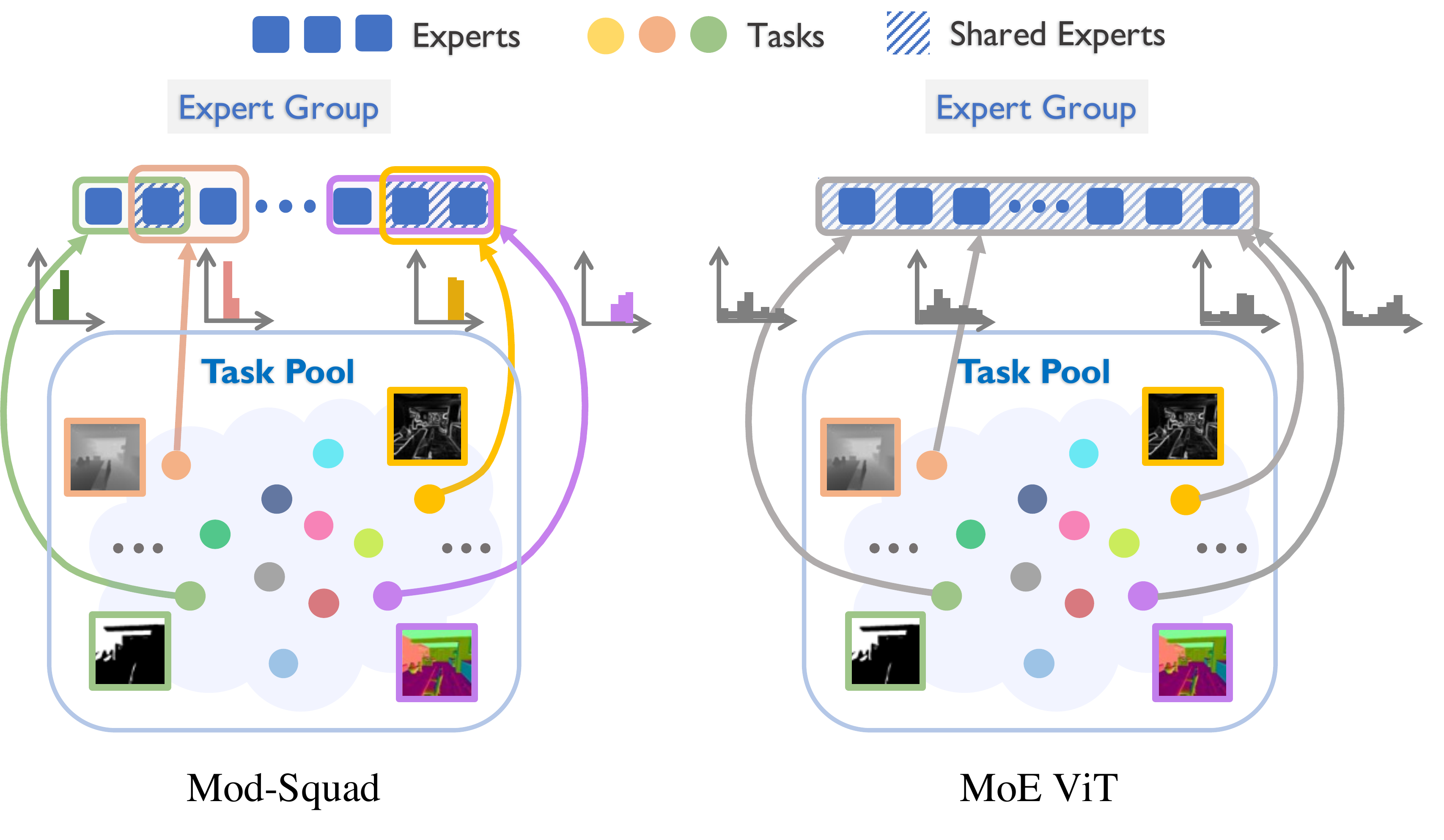}
  \caption{\textbf{A comparison between \model and MoE ViT.} Our key motivation is that experts should
leverage commonalities in some tasks (cooperation) but focus on a subset of tasks that require specific features and do not interfere with each other (specialization).
  }
  \label{fig:motivation}
\end{figure}
Computer vision involves a great number of tasks including recognition, depth estimation, edge detection, etc. Some of them have a clear and strong relationship: {they are likely to benefit from shared features. An example would be a task to classify cars and pedestrians and a task to segment the same classes. Other tasks appear to be less related: it is not clear what features they would share. An example could be tumor detection in medical images and face recognition.} 

Multi-task learning (MTL) aims to model the relationships among tasks and build a unified model for a diverse set of tasks. On the one hand, tasks often benefit by sharing parameters, i.e., {\bf cooperation}. On the other hand, some tasks may require specialized expertise that only benefits that single task, i.e., {\bf specialization}.
%As an example, consider an autonomous driving system that must do pedestrian detection, distance estimation, scene recognition and other diverse tasks simultaneously. The system, while benefiting from learning all of these tasks, may also require some special expertise for each of them to achieve high accuracy. 
A good MTL system should be flexible to optimize experts for the dual purposes of cooperation and specialization.
% allocate parts of the model for sharing knowledge among tasks and other parts for the specialized expertise needed in a single task. 
% \e{The following sentence really describes 4 things, not 2: 1) flexibility (to do what?), 2) accuracy, 3) low computation, 4) low parameter count. Obviously, numbers 3 and 4 are related, but they are not the same. I'm not sure what you're really trying to say here.} 
% Flexibility to optimize experts for the dual purposes of cooperation and specialization and efficacy in accuracy, computation cost, and parameters size are two main targets for a well-designed MTL system.
% \yk{I am not sure if the autonomous driving system is a good example. Our system can't perform different tasks simultaneously, and the example is not related to previous statements about shared and specialized knowledge.} 
% An ideal MTL system can combine the best of two worlds including \textbf{efficacy} and \textbf{flexibility}. 

There are two well-known challenges in MTL: (1)~gradient conflicts across tasks~\cite{chen2020just, yu2020gradient}; and (2)~how to design architectures that have both high accuracy and computational efficiency.
% efficient architecture that have high accuracy for all tasks while having short training and inference times. 
% \e{Again, in the previous sentence, I'm worried that the word 'efficient' is too vague. You could say something like, 'how to design architectures that have high accuracy while having short training and inference times', or something like that..... It's too vague right now.}
% efficient (in both computation and capacity) architecture for all tasks. 
% \e{I'm not sure exactly what you mean by `capacity'. Is that equivalent to memory footprint? Or are you focused on the number of parameters?  Also, for computation, are you concerned more with training time or inference time? Once I know exactly what you want to emphasize, I think we can reword it somewhat.} \zt{Compuation cost for single forward.}
% \e{Not sure what you mean by automation here.} 
% the flexibility to adjust the model in both structure and parameters according to the given task\cite{zhang2021automtl}.
Previous efforts include manually designing architectures \cite{caruana1997multitask} or conducting neural architecture search \cite{ahn2019deep} to induce cooperation and specialization in different parts of the model. However, these methods either require heavy manual customization, reducing generality and limiting applicability, or require very long training times. 
%Further, these implementations are based on specific model structures, making them hard to generalize and to benefit from more advanced network architectures (\eg transformers).
% Furthermore, it could be hard to generalized to all kinds of tasks.
% \e{The next sentence, in my opinion, is a bit vague. It `could be'?  What does that mean? Are you trying to say that these models have been unable to demonstrate generalization? Also, what do you mean by `all kinds of tasks'? I think you need to be more specific here.} 

% \e{I think here, you want to say something about what is lacking in other models...}

% \e{I suggest the following start to the next section. See if you agree.}
To address these challenges, we introduce {\bf Mod-Squad}, a new model that constructs a Mixture of Experts (MoE)~\cite{shazeer2017} to be {\bf mod}ularized multi-task learners (a {\bf squad}). {
% \yk{Different from previous MoE-based methods (e.g., MoE ViT~\cite{riquelme2021scaling}),} 
Our design allows experts to cooperate on tasks {\bf when it is helpful}, rather than penalizing experts that do not participate in {\em every} task. At the same time, some experts naturally develop a deep specialization in particular tasks, improving performance.} 
{The left figure in Fig.~\ref{fig:motivation} shows an example of the specialization and cooperation of experts in \model.}
A further and important side benefit, discussed below, is that this sparsification of experts allows our model to be decomposed into {much smaller single-task models that perform extremely well}.
% pieces that perform extremely well separately.

We achieve these goals by first integrating mixture of experts (MoE) layers into our vision transformer~\cite{dosovitskiy2021an} backbone network. 
% as shown in Fig.~\ref{fig:pipeline}. 
The motivation is to divide the model into groups of experts, and for each expert to construct a minimum part of the model that can be shared among tasks or be specialized for one task. The experts can have any network structure (e.g., MLP or attention network~\cite{zhang2022mixture}) so that we can incorporate advanced model designs.
% A task-dependent routing network is employed at every MoE layer to select experts for each input token. For every expert, its activation (or deactivation) depends on the task and the input.
Our modular design allows cooperation and specialization via the distribution of tasks to experts and also experts to tasks.  Below, we formalize this idea mathematically by analyzing the probability distribution over tasks and experts, and using a novel loss function to induce a specific structure on this distribution. 
%We first briefly introduce the limit of previous MoE work and then explain the motivation and strength of our novel loss.

% \cite{liang2022m, mustafa2022multimodal}
Many previous MoE works~\cite{shazeer2017, riquelme2021scaling, zhang2022mixture} use a load-balancing loss that encourages the frequency of expert usage (across all tasks and batches) to be highly similar. Some MoE methods~\cite{liang2022m, mustafa2022multimodal} directly apply this loss after the forward pass of each task on the multi-task scenario so that each task evenly uses all experts.
However, this approach may force experts to set parameters on conflicting tasks with learning gradients that counteract each other. In other words, while an expert may benefit from being shared among certain pairs of tasks, it may be harmed by being forced to share among other pairs of tasks. This is an explanation for the difficulty of training multi-task models under such an expert-balancing loss.
% However, this ignore the fact that experts could be interfered by some tasks when trying to leverage commonalities in all tasks and make training much harder when target on more tasks with complicated relation. 

In comparison, we contend that experts should leverage commonalities in some tasks (cooperation) but also create a subset of experts that learn specific features (as needed by some tasks) and do not interfere with each other (specialization). Such an assignment of tasks to experts can be represented via \textbf{a sparse but strong dependence between experts and tasks}. Fig.~\ref{fig:motivation} illustrates this key difference between our model and previous MoE work, showing how our model induces a sparser structure in the assignment of experts to tasks. 
% \gc{remove refering to figure 5}
% (1)\textbf{experts are shared only among related tasks}; (2)\textbf{the assimilation and specialization pattern should be learned automatically}.
To implement this idea, we add a loss term to maximize the mutual information between experts and tasks. This induces a strong dependency between experts and tasks, with each task heavily related to a small set of experts and vice versa. 
% Intuitively, this enables the model to dynamically self-organize with respect to sharing and specialization. It helps address the gradient conflicts problem and improve some degree of freedom in architecture as each task is allocated with a different numbers of experts.

% \e{It also improves ... as well as the capacity and some degree of freedom in architecture.  Do you have evidence of this? Can you be more clear about exactly what you are trying to say?}
% Furthermore, each experts only allow gradients from the chosen tasks thus addressing the gradient conflicts challenge. The dynamically self-organize property enable automation for each task and give the model great flexibility.

% It is critical to assign experts only a subset instead of all the tasks especially when we scaling up multi-task learning and each expert contribute to over 10 different tasks. In that case, experts suffer from no specialization and low capacity for each task with the overwhelming tasks number.
% and tasks and optimize the joint distribution over tasks and experts pairs. 
% As a result, strong dependence between tasks and experts are developed as well as two important property: balance and sparsity. 

Interestingly, we find that our model converges to a state in which, after training, most experts are never or rarely used for many tasks (evidence of specialization),  but the experts are still balanced in their activation frequency. This property enables us to extract a compact sub-network from the giant model for each task. The small networks extracted in this fashion work independently as standalone models for individual tasks with {\em no performance drop}. This property enables us to train a giant, sparse model in a scaled-up multi-task learning scenario and later get compact sub-networks for each task with high performance. 

\noindent Our main contributions can be summarized as follows:

\vspace{-2mm}
\begin{itemize}[align=right,itemindent=0em,labelsep=2pt,labelwidth=1em,leftmargin=*,itemsep=0em]

\item \textbf{Modular multi-task learner.} We propose a new modular backbone model, {\model}, that is composed of a large group of attention and feed-forward experts. The experts can be flexibly assigned a subset of tasks to achieve specialization and cooperation. 
% The experts are assigned to a subset of the tasks instead of all tasks to address gradient conflict among tasks.
% \yk{We propose a new modular backbone model, \textbf{\model}, that is composed of a large group of attention and feed-forward experts. It sparsely activate different experts condition on the task and input.}

\item \textbf{Optimizing the joint distribution over tasks and experts}.
%\yk{Optimizing the mutual dependency between tasks and experts?}} 
\model includes a new loss term that encourages a sparse but strong dependence between experts and tasks. This is done by measuring and maximizing the mutual information between tasks and experts. 
% encourages each task to consistently activate a subset of the experts. 
% \yk{Based on \model, we propose a new auxiliary loss that encourages each task to consistently activate a subset of the experts.}

\item \textbf{Effective and Efficient multi-task learners at scale.} Experiment results show that \model achieves state-of-the-art performance on two major multi-task datasets while maintaining its computational efficiency. 
% \yk{Experiment results shows that \model achieves state-of-the-art performance on two major multitask datasets.}

\item \textbf{Extracting small sets of experts as standalone models with no performance drop.} We further show that \model can be effectively pruned for a designated task without sacrificing performance.
% \yk{We further show that the large multitask model can be effectively pruned for a designated task without sacrificing the performance.}

% \item design MOE and new loss for multi-task learner
% \item Find the pruning works.

% \item Strong results

\end{itemize}

\section{Related Work}
\noindent \textbf{Multi-task Learning.} Multi-task learning jointly learns multiple tasks by sharing parameters among tasks. One common approach is to manually design the architecture, sharing the bottom layers of a model across tasks \cite{caruana1997multitask, kokkinos2017ubernet, bragman2019stochastic}. Some works~\cite{vandenhende2019branched} design the architecture according to task affinity. Others~\cite{ahn2019deep, bragman2019stochastic, sun2020adashare} leverage Neural Architecture Search or a routing network~\cite{rosenbaum2018routing} to learn sharing patterns across tasks and automatically learn the architecture. Recently, transformer-based MTL architectures~\cite{xumtformer}  have been explored and have shown advantages over CNN-based models. 
In comparison, we customize MoE layers into vision transformers; each MoE module constructs a minimum part of the model that can be distributed to a subset of all tasks instead of all tasks. As a result, our model is flexible in its creation of cooperation and specialization.
% Compared to these work, our transformer-based approach automatically learn to share parameters among tasks and can adapt the structure and capacity in regard to different tasks. 

\noindent \textbf{Mixture of Experts (MoE). } The MoE was first proposed by Jacobs et al.~\cite{jacobs1991adaptive} as a technique to combine a series of sub-models and perform conditional computation. Recent work~\cite{shazeer2017} in NLP proposes sparse MoE to reduce computation cost, and some works~\cite{lepikhin2021gshard, JMLR:v23:21-0998} train gigantic models with trillions of parameters based on the sparse model. Some have used the MoE technique to train huge models in vision~\cite{riquelme2021scaling, wu2022residual} or multi-modal applications~\cite{mustafa2022multimodal}. 
% \e{When you say 'this technique', do you mean using a 'sparse model'. Does that undercut our contribution?} 
% More recently, M$^3$ViT~\cite{liang2022m} uses MoE techniques to design a multi-task learning model that is computationally efficient during  training. \gc{move this sentence later}
These works typically focused on combining the Feed-Forward Network layer with the MoE or develop a better routing strategy~\cite{lewis2021base, nie2021dense}. 
MoA~\cite{zhang2022mixture} proposes a new module that combines the attention network with the MoE while having a low computational cost and the same parameter budget as a regular attention network. 
More recently, M$^3$ViT~\cite{liang2022m} uses MoE techniques to design a multi-task learning model that is computationally efficient during  training.
Compared to these previous methods, we demonstrate a MoE model that is not only computationally efficient, but is also flexible as a modularized multi-task learner that can easily induce both cooperation and specialization. Although M$^3$ViT~\cite{liang2022m} also use MoE in their approach, the experts in their model share between all tasks and cannot be specialized for tasks. 

% Compared to all other MoE work including M$^3$ViT\cite{liang2022m}, we show that the MoE can not only be computation efficient, but also a flexible modular design that can quickly adapt in keeping with the requirement of tasks. 
% Compared to M$^3$ViT\cite{liang2022m}, beside the technical difference and contribution, we also scale up multi-task learning and demonstrate that experts can be specialize for some tasks instead of being used by all tasks, which give higher performance when tasks number grow up and more parameter-efficient when doing single task inference. 

\noindent \textbf{Pruning.} {Pruning refers to the process of removing components of a larger model to produce a smaller model for inference, with the goal of maintaining as much accuracy as possible while improving runtime computation efficiency.} Generally, pruning is categorized into \textit{unstructured pruning}~\cite{han2015deep_compression}, which removes individual weights that have a minimal contribution to accuracy and \textit{structured pruning}~\cite{he2018soft, li2019learning}, which ranks filters or blocks and prunes these based on some criterion. Usually, extra fine-tuning is conducted for the pruned network to help maintain the performance~\cite{Renda2020Comparing, liu2018rethinking, yu2018slimmable}. 
% pruning and training need to be performed  iteratively and repeatedly~\cite{Renda2020Comparing, liu2018rethinking, yu2018slimmable}. 
% \e{Hmmm... the previous sentence seems to contradict my definition that pruning happens after model training. I'm not famililar with iteratively pruning and training. What's the advantage of that over just training a smaller model in the first place? I suggest that you try to define pruning in the first sentence of this paragraph so that it is consistent with how people use it in the citations here.} 
Most of pruning is for single task and very few of them consider the case in multi-task learning. 
In this work, our proposed model has a unique property that a series of small sub-network for each task can be extracted from it with no performance drop and no additional fine-tuning. This is somehow similar to pruning but more likely to be an advantage of our model rather than a new way of pruning. 
% \zt{Check the last sentence.}
% In this work, we propose a new way of pruning that can extract a series of sub-network for each task from a large multi-task model with no performance drop and additional training. 
% \zt{Do we need to compare with some other pruning method that can prune our big multi-task model?} 

% Our approach is similar to pruning as we both can remove some parameters from the model. The major difference is we focus on MTL and we only need to train then can get a pruned model for each task. 
\begin{figure*}
  \centering
  \includegraphics[width=17cm]{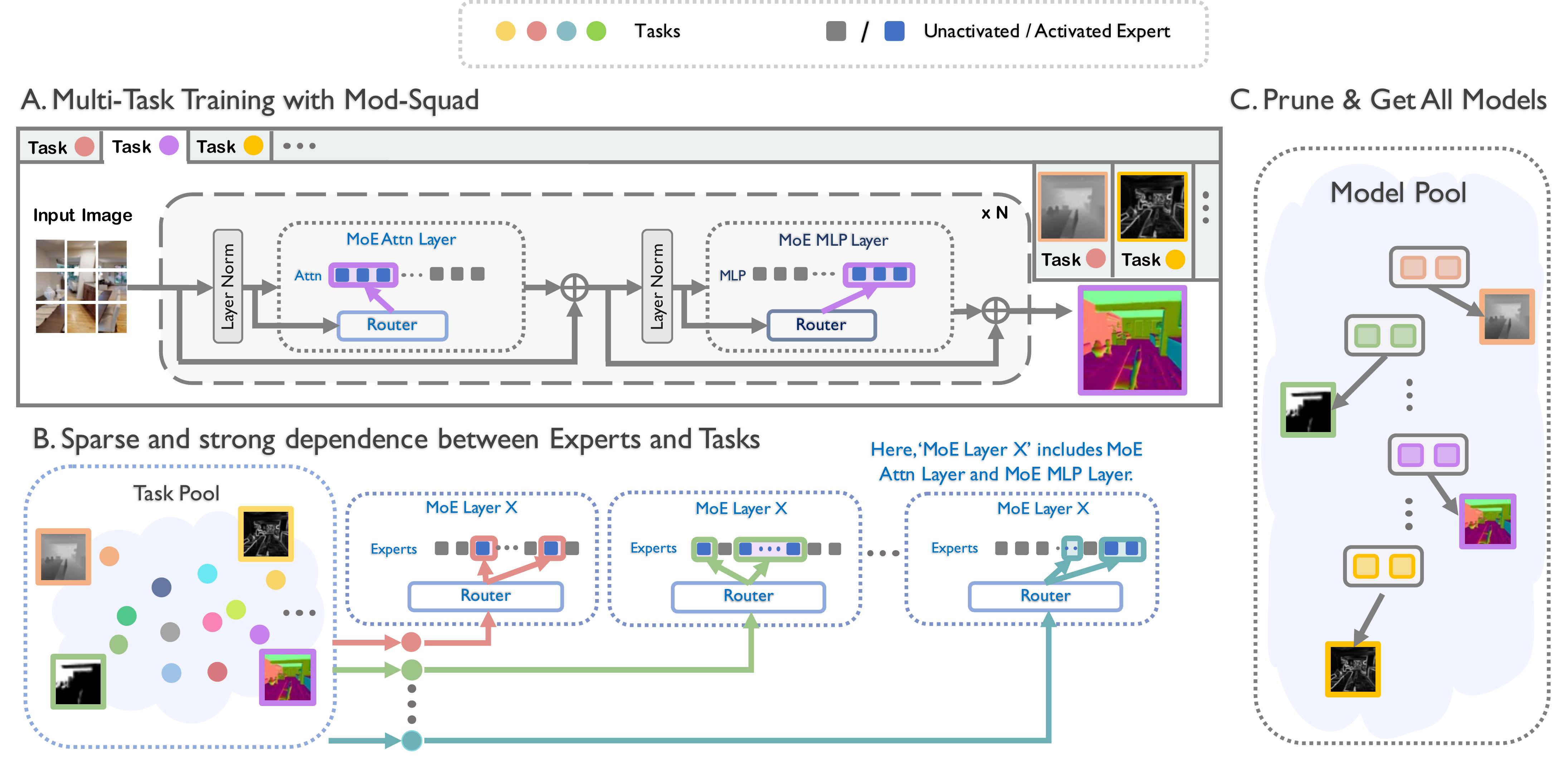}
  \caption{\textbf{The pipeline of our multi-task foundation model.} 
  Each transformer block in \model consists a MoE attention network (MoE attn.) and a MoE MLP network. The multi-task model \model is trained with our proposed mutual information loss. \model develops a strong dependence between experts and tasks. Then we can extract a small sub-network from \model for each task with no performance drop.
%   For each task, we can extract a small sub-network after training with no performance drop compared to the large multi-task model. 
%   \gc{Needs to be updated}
  } 
  \label{fig:pipeline}
\end{figure*}

\section{Method}

We start with the definition of multi-task learning.
Suppose we have $M$ tasks $T_1, T_2, ...,T_M$ and $Q$ images $I_1, I_2, ..., I_Q$. We define a task $T$ as a function that maps image $I_q$ to $T(I_q)$. 
% \e{Here, if you are defining the function as $G_T$, don't use $G_{T_i}$ in the definition. Use $G_T$. Later, when there is more than one task, you can use $G_{T_i}$.}. 
Our dataset $D$ contains for each task $T_i$ a set of training pairs $(I_q; T_i(I_q))$, e.g. (image; depthMap). Here, for simplicity, we assume that every task contains a training pair for every one of the $Q$ images, but note that our approach can be extended to the case in which every task contains a different subset of images in its training pairs.

\subsection{Preliminaries}

\noindent\textbf{Mixture of Experts.} A Mixture of Experts (MoE) layer typically contains a set of expert networks $E_1, E_2, ..., E_N$ along with a routing network $G$. The output of a MoE layer is the weighted sum of the output $E_k(x)$ from every expert. The routing network model $G$ calculates the weight $G^k$ for each expert given input $x$. Formally, the output of a MoE layer is 
% \e{In the following expression, you put $k$ after the parentheses. This is highly unusual notation. Do other people do this? If so, you should at least explain it. Or, perhaps better, use $G^k(x)$ meaning the $k$th output of $G$, or something like that. Make sure you define this notation, especially since it appears we have plenty of room.}
\begin{align} \label{eqn:moe_output}
y =&\sum_{k=1}^{N} G^k(x)  E_{k}({x}).
\end{align}

The routing network $G$ is a Noisy Top-$K$ Routing network \cite{shazeer2017}
% \e{If the reader is supposed to already know what a `Noisy Top-K Routing network' is, then you should at least have a citation to it. If you are introducing the concept (which I think you are), I would use language more like, 'We refer to $G$ as a Noisy Top-K Routing Network, which means that it ....'.} 
with parameters $W_g$ and $W_{noise}$. It models $P(E_k|x)$ as the
probability of using expert $E_k$ and selects the Top-$K$ to contribute to the final output. The whole process is shown as follows:
\begin{align} \label{eqn:moe_output}
    % G(x) =&  \operatorname{TopK}(P(E|x)) \\
    % P(E |x) 
    G(x) =&  \operatorname{TopK}(\operatorname{Softmax} (xW_g  \nonumber \\
    & + \mathcal{N}(0,1) \operatorname{Softplus}(xW_{noise}))),
\end{align}
where $\operatorname{TopK}(\cdot,k)$ sets all elements in the vector to zero except the elements with the largest $K$ values, $\operatorname{Softplus}$ is the smooth approximation to the ReLU function:
\begin{align}
    \operatorname{Softplus}(x) =& log \left( 1+\exp \left( x \right) \right). %\\
    % \operatorname{TopK}(v,K)=& \left\{\begin{array}{ll}
    % v_i & \text{if $v_i$ is in the top $K$ elements} \\
    % 0 & \text{otherwise}
    % \end{array}\right.
\end{align}

\subsection{\model} % Modular design
\model is a multi-task model with the vision transformer as the backbone network and several parallel task-specific heads. 
As shown in Fig.~\ref{fig:pipeline}, a key design in our model is customizing MoE into the vision transformer so that each expert can construct a minimum part of the model that can be either shared between tasks or specialized for tasks. 
% However, different from most previous MoE work \cite{JMLR:v23:21-0998, riquelme2021scaling} that focus on computation efficiency, our motivation is to build a more flexible and controllable architecture. 
% The expert and gating function enable the model to choose experts according to the task. 
Specifically, we customize the MoE attention block (MoA)~\cite{zhang2022mixture} and MoE MLP block~\cite{shazeer2017} into the transformer layer. 
Each MoE block consists of $N$ experts $E_1, E_2, ..., E_N$ which can be either an attention head or an MLP layer along with $M$ \textbf{task-specific routing networks} $G_1, G_2, ..., G_M$ that select experts conditioned on input tokens.
Note that each routing network $G_i$ has its own parameters $\left( W_g^i, W_{noise}^i \right)$. 
We also add a learnable task embedding to the hidden input state so that each expert is aware of the target task. Thus, in \model, the output of each MoE layer is
\begin{equation}
    y = \sum^N_{k=1} G^k_i(x) \cdot E_k \left( x + e_i \right),
\end{equation}
where $i$ is the task id and $e_i$ is the respective task embedding.

% \e{Zitian, let me try adding a subsection here to see if it clarifies some issues. We can delete it later if it is not right.}

\subsection{A joint probability model over tasks and experts} 
In order to model cooperation and specialization, we define a probability model over tasks $T$ and experts $E$. We assume that when our trained network is deployed, it will be assigned a random task $T$ according to a global distribution over tasks $P(T)$. (Typically we assume this distribution to be uniform over tasks.) Subsequently, it will be given a random image $X$ according to $P(X|T)$.

For a given MoE layer, we model the probability $P(E_i|T_j)$ of using expert $E_i$ with task $T_j$ as the frequency with which $E_i$ is assigned to task $T_j$ by the routing network. For example, for 100 images in task $T_j$, if the routing network assigns 30 of them to expert $E_i$, then $P(E_i|T_j)=0.3$. Since the routing network does not make hard assignments of experts to tasks, but rather assigns weights resulting from a softmax function to each expert, we sum these soft weights to measure the frequency: 
%\zt{I change $G(E_i|T_j)$ to $P(E_i|x_k)$}
$$
P(E_i|T_j)=\sum_{k=1}^{Q_{T_i}} G_{T_j}^{E_i}(x_k),
$$
%$$
%P(E_i|T_j)=\sum_{k=1}^{N_{T_i}} P(E_i|x_k),
%$$
where $G_{T_j}^{E_i}$ gives the weight for expert $E_i$ for task $T_j$ on the input $x_k$ from image $I_k$. $Q_{T_i}$ is the number of images for task $T_i$.
% and $P(E_i|x_k)$ is the probability output by the routing network.
Given this definition of conditional probability, the joint probability $P(E,T)=P(E|T)P(T)$, and of course, we can obtain $P(E)=\sum_T P(E,T)$.

A key intuition in our work is that {\bf experts should be dependent on tasks}, that is, experts should specialize in specific tasks, at least to some extent. This notion can be captured by measuring the {\em mutual information (MI)} between tasks and experts, using the probability model defined above:
\begin{align} \label{eqn:MI}
I(T;E) =& \sum_{i=1}^M \sum_{j=1}^N P(T_i,E_j) \log \frac{P(T_i,E_j)}{P(T_i)P(E_j)}.
\end{align}
If experts are assigned with equal frequency to all tasks, then the mutual information will be 0. If each expert is assigned to exactly one task (when $M=N$), then the dependence (and hence the mutual information) is maximized.

\subsection{Maximize mutual information between experts and tasks}

% Our key motivation is that cooperation and specialization can be formalized as the process of matching experts and tasks. 
% {Thus, it is critical to assign experts only a subset of all the tasks especially when we increase the number of tasks.
% Such that experts can avoid the gradient conflict problem and capture shared knowledge between tasks.}
% We address this challenge by introducing a new loss to maximize the mutual information between experts and tasks, which encourages specialization. The new loss is combined with the multi-task training loss which naturally prefers cooperation to benefit from shared features. 

% {To strengthen the dependence among experts and tasks, we optimize their mutual information:}
% \begin{align} \label{eqn:MI}
% I(T;E) =& \sum_{i=1}^M \sum_{j=1}^N P(T_i,E_j) \log \frac{P(T_i,E_j)}{P(T_i)P(E_j)}
% \end{align}
% where $P(x_k, T_i)$ is the probability of a example $(x_k, T_i)$, $P(T_i)$ is the probability that a random example belong task $T_i$. We can further define $P(T_i,E_j)$ and $P(E_j)$ as follows:
% \begin{align}
% P(T_i,E_j) =& \sum_{k=1}^N P(x_k, T_i)P(E_j|x_k,T_i) \\
% P(E_j) =& \sum_{k=1}^N \sum_{i=1}^M P(x_k, T_i)P(E_j|x_k,T_i) 
%     %   =& \sum_{i=1}^M \sum_{j=1}^K \sum_{k=1}^N P(E_j|x_k,T_i) \log \frac{\sum_{k=1}^N P(E_j|x_k,T_i)}{P(T_i)P(E_j)} 
% \end{align}
% {where $P(E_j|x_k, T_i)$ is the probability of selecting expert $E_j$ modeled by the routing network and $x_k$ is the input token from the $k$-th training samples.}

To understand what mutual information do, we break down the Equation. \ref{eqn:MI} as following:
\begin{align} \label{eqn:MI_split}
I(T;E) =& \sum_{i=1}^M \sum_{j=1}^K P(T_i,E_j) \log P(T_i,E_j) \nonumber \\
& - \sum_{i=1}^M P(T_i)  \log P(T_i) \nonumber \\ 
& - \sum_{j=1}^K P(E_j) \log P(E_j).
\end{align}

% Here as claim before we assume that every tasks are trained on same amounts of data and share the training examples for simplicity. But the mutual information can still be calculated even if each task has its own data. So in this assumption, we have $P(T_i) = \frac{1}{M}$ and $P(x_k, T_i) = \frac{1}{MN} $ \yk{Is this a important assumption? I think we don't need this assumption to make the model work.}

% In Equation. \ref{eqn:MI_split}, the first term is the entropy of $P(T_i,E_j)$ encourage sharp distribution of $P(T_i,E_j)$. The distribution prefer to be either as large as possible or just zero. 
% The second term is the entropy of $P(T_i)$ which is a constant and can be ignored. The third term is the minus entropy of $P(E_j)$ and encourage smooth distribution of $P(E_j)$, so the experts prefer to have the same used frequency. 

In Eq.~\ref{eqn:MI_split}, the first term is the negative entropy of $P(T_i,E_j)=P(E_i|T_j) P(T_j)$. Maximizing this term encourages the sharpness of the conditional distributions $P(E_i|T_j)$, since $P(T_j)$ is a constant decided by data distribution, and is not affected by model parameters. 
% In other words, the distribution is encouraged to be either as large as possible or just zero. 
The second term is the entropy of $P(T_i)$ which, again, is a constant and can be ignored. 
The third term is the entropy of $P(E_j)$. Maximizing the term encourages a high-entropy or flat distribution of $P(E_j)$, encouraging the experts to be evenly used across the entire dataset. 

In practice, we add $-I(T;E_Y)$ to our total loss for each MoE layer $Y$ with a weight parameter $w_{MI}$ where $E_Y$ represents all the experts in $Y$. 
% The total loss function for our model is the mutual information loss with the weighted sum of loss from each task. 
We follow \cite{kendall2018multi} to learn an auto-balancing weight $w_T$ for each task $T$ and add the task-specific loss $L_T$ for all tasks. So the total loss is
\begin{equation}
    \mathtt{L} = \sum^M_{i=1} w_{T_i}L_{T_i} - w_{MI} \sum_{\forall MoE \ \mathtt{layers}\ Y} I(T;E_Y).
\end{equation}

\subsection{Train Once and Get All}

\label{sec:pruning}

% A unique property in out model is that only a small subset of experts for each task are used for each task. 
In previous MoE works\cite{liang2022m, mustafa2022multimodal}, they use a subset of the experts for one input image but all the experts for  each task. In comparison, \model activates a subset of the experts when forwarding both single image and multiple images from the same task. 
% they use sparse model for a single image but compact and dense model for a single task.
% The entire model are activated during inference time or further fine-tuning for each task.
% % \yk{but they always use the entire model for each task}. 
% In comparison, our model is sparse for both image-level and task-level. 
% Further, all experts in our model are used in the same level of frequency for the whole multi-task dataset, which guarantee the efficiency of our model. 
Further, all the experts are evenly used in \model when forwarding the whole multi-task dataset. This guarantees that the capacity of \model is fully utilized and not wasted.
A typical relation between tasks and experts will be demonstrated in \cref{relation}.
% A typical relation between tasks and experts in one of the MoE layers can be seen in Fig.~\ref{fig:expert_task}.\gc{check if figure correct}

Benefiting from the constant sparsity of \model at the image-level and the task-level, unused or rarely used experts can be removed in each MoE module when doing single-task inference. This can be done by counting the using frequency of each expert for the task and removing those experts with smaller frequency than a threshold $\theta$. Note that some tasks could use more experts and others use less for each MoE layer. For example, a low-level task may require more experts at the first few layers of the network and a high-level task may require more experts at the last few layers of the network. \model is capable of dynamically self-organize architecture and selecting experts according to the requirement of tasks, which provides some degree of freedom in architecture and extra flexibility in allocating model capacity.
% This meet the requirement of flexibility of an ideal MTL system.

% Intuitively, this enables the model to dynamically self-organize with respect to sharing and specialization. It helps address the gradient conflicts problem and improve some degree of freedom in architecture as each task is allocated with a different numbers of experts.

After removing experts, our pruned model can be directly deployed for the respective task. Since the removed experts are never or rarely used, the pruned model achieves the same level of performance as the original model but with a much smaller number of parameters and without any fine-tuning. In the case where we set $\theta=0$ and keep all the experts that have ever been used, we observe no drop in performance {while still effectively pruning a large portion of the model}. This removing experts process is similar to pruning, but we just adapt a simple thresh then remove strategy and no additional training is needed like in some of the pruning work\cite{cai2020once}. Once training, a series of small sub-networks can be extracted for all tasks. This property enables us to build a very large model benefit from all tasks, but only requires a fraction of model capacity for single-task inference or fine-tuning.
% We will demonstrate this in experiments in the next section.

% Compared to previous MoE work that use all experts for each task, we only use a small subset of experts for each task but all the experts in the same level frequency for all tasks. We refer to this as \textbf{sparsity in single task} and \textbf{density in all tasks}. 

% As stated before, efficacy and flexibility are two important aspects in MTL. 
% Due to the modular design of MoE and the sparsity in single task, only a small portion of experts in each transformer block are being used for each task. Some layers may have more experts while other may have less. This phenomenon are shown in Fig. \ref{fig:task}. 

% As a result, when conducting single task inference, we can remove a large portion (60\%-70\%) of experts that are rared or never been used and still keep almost the same performance. This is similar to pruning \cite{han2015deep_compression} but we do not need any fine-tuning. This is also similar to neural architecture search \cite{zoph2017neural} as each transformer block has different shape and size. However, we only require training one model once instead of training multiple models with different architecture.

% Why is mutual loss superior?
% Neural network implicitly conduct the assimilation and specialization for every part in the model. 

\section{Experiment}

% A set of experiments are conducted to demonstrate the superiority of our model compared to several competing baselines and other state-of-the-art MoE approaches. We make an apples to apples comparison in terms of task performance, model size, and computation cost. \yk{This paragraph can be removed if need to save some space}

\begin{table*}[t]

\begin{center}
\footnotesize
\tabcolsep=0.07cm
\begin{tabular}{c|c|c|c|cc|ccc|c|c|c|c|c|c}
\toprule
\multirow{3}{*}{Model} & \multicolumn{1}{c|}{Obj. Cls.}  & \multicolumn{1}{c|}{Scene Cls.}    & \multicolumn{6}{c|}{Depth Euc.}     & \multicolumn{1}{c|}{Normal} & \multicolumn{1}{c|}{Curvature} & \multicolumn{1}{c|}{Reshading} & \multicolumn{1}{c|}{Edge3D} & \multicolumn{1}{c|}{Keyp.2D} & \multicolumn{1}{c}{Segm.2D}\\ \cline{2-15}
                    & \multirow{2}{*}{$Acc(\%) \uparrow$} & \multirow{2}{*}{$Acc(\%) \uparrow$} & \multirow{2}{*}{$RMSE \downarrow$} & \multicolumn{2}{c|}{Error $\downarrow$} & \multicolumn{3}{c|}{$\delta$, within $\uparrow$} &    \multirow{2}{*}{L1 dis. $\downarrow$} &    \multirow{2}{*}{L2 dis. $\downarrow$}  &    \multirow{2}{*}{L1 dis. $\downarrow$} &    \multirow{2}{*}{L1 dis. $\downarrow$} &    \multirow{2}{*}{L1 dis. $\downarrow$} &    \multirow{2}{*}{L1 dis. $\downarrow$}             \\ \cline{5-9}
                                    &          &         & &  Abs.  &     Rel.   &   1.25 & $1.25^2$ & $1.25^3$   &    & & & & &                                 \\
\midrule                     

STL & 56.5 & 60.0 & 6.94 & 0.089 & 1.77 & 92.8 & 96.9 & 98.7 & 0.403 & 1.12 & 0.184 & \best{0.119} & 0.0312 & 0.171\\
MTL & 57.3 & 64.9 & 6.75 & 0.084 & 1.26 & 93.0 & 97.0& 98.9 & 0.386 & 1.06 & 0.170 & 0.127 & 0.0284 & 0.166\\
$M^3ViT$\cite{liang2022m}  & 58.0 & 65.6 & 6.69 & 0.083 & 1.26 & 93.2 & \best{97.2} & 98.9 & 0.383 & 1.05 & 0.174 & 0.126 & 0.0289 & 0.164 \\
\hline
\model & \best{59.0} & \best{66.8} & \best{6.59} & \best{0.082} & \best{1.25} & \best{93.3} & \best{97.2} & \best{99.0} & \best{0.374} & \best{1.02} & \best{0.167} & {0.123} & \best{0.0275} & \best{0.161}  \\
% Ours-Pruning &  & & & & & & & & & & & & &  \\

\bottomrule
\end{tabular}
\end{center}
\caption{\textbf{Metric for each task on the taskonomy dataset.} For each task, we use different metrics to evaluate its performance. More results on other tasks can be found in the supplementary.} 
\label{tab:metric}
\end{table*}

\begin{table}[]
    \centering
    \small
    % \hspace{-0.5in}
    \setlength{\tabcolsep}{1mm}{
    \begin{tabular}{l|c|cc|cccc} %>{\columncolor{mygray}}c
    \hline
    Method & STL & MTL & M$^{3}$ViT & MLP & Attn & Ours & Pruning \tabularnewline
     \hline
    % Params(M) & 86.4 & 90.0 & 176.4 & 176.4 & 105.6 & 190.5 & 99.2 \tabularnewline
    Params(M) & \best{86.4} & 90.0 & 176.4 & 176.4 & 105.6 & 201.3 & 116.9 \tabularnewline
    FLOPs(G) & \best{17.7} & 18.5 & 19.7 & 19.7 & 19.7 & 19.7 & 18.4  \tabularnewline
    \hline \hline
    Object Cls.& 0.0 & +1.4 & +2.6 & +3.0 & +3.0 & \best{+4.4} & \best{+4.4} \tabularnewline
    Scene Cls. & 0.0 & +8.1 & +9.3 & +10.0& +9.6 & \best{+11.3}& \best{+11.3}\tabularnewline
    Depth Euc. & 0.0 & +2.7 & +3.6 & +3.9 & +4.4 & \best{+5.0} & \best{+5.0} \tabularnewline
    Depth Zbu. & 0.0 & +2.1 & +2.4 & +2.6 & +2.4 & \best{+2.8} & \best{+2.8}\tabularnewline
    Normal     & 0.0 & +3.5 & +4.2 & +4.5 & +4.5 & \best{+6.5} & \best{+6.5}\tabularnewline 
    Curvature  & 0.0 & +5.3 & +6.2 & +7.1 & +6.2 & \best{+8.9} & \best{+8.9} \tabularnewline 
    Reshading  & 0.0 & +7.6 & +5.4 & +5.9 & +8.1 & \best{+9.2} & \best{+9.2} \tabularnewline 
    Edge2D     & 0.0 & +0.6 & +2.0 & +1.8 & +1.2 & \best{+3.6} & \best{+3.6} \tabularnewline 
    Edge3D     & \best{0.0} & -6.7 & -5.8 & -4.2 & -5.8 & -3.3 & -3.3\tabularnewline 
    Keyp.2D    & 0.0 & +5.3 & +3.6 & +3.6 & +6.3 & \best{+8.3} & \best{+8.3}\tabularnewline 
    Keyp.3D    & 0.0 & +1.3 & +2.7 & +4.1 & +2.7 & \best{+5.5} & \best{+5.5}\tabularnewline
    Segm. 2D.  & 0.0 & +2.9 & +4.0 & +5.2 & +3.5 & \best{+5.8} & \best{+5.8} \tabularnewline 
    Segm. 2.5D & 0.0 & +1.9 & +3.2 & +3.8 & +3.2 & \best{+5.1} & \best{+5.1} \tabularnewline 
    
    \hline
    \rowcolor{mygray} Mean & 0.0 & +2.8 & +3.3 & +3.9 & +3.8 & \best{+5.6} & \best{+5.6}  \tabularnewline
    % Scene Cls. &  & & & & & & \tabularnewline
    % Depth Euc. &  & & & & & & \tabularnewline
    % Depth Zbu.  &  & & & & & & \tabularnewline
    % Normals  &  & & & & & & \tabularnewline
    % Curvature  &  & & & & & & \tabularnewline
    
    \end{tabular}
    }
    \caption{ \textbf{Comparison of $\Delta_t$ between MTL methods on the Taskonomy.} We report their average drop for each task with respect to the vanilla single-task model. MLP and Attn represent using only MoE MLP and MoE attention network in the backbone respectively. }
    \label{tab:MTL}
    % \vspace{-0.2cm}
\end{table}

\subsection{Experiments Settings}

% [02:15:41.919489] len train:  3793769
% [02:15:41.919520] len val:  60454*10
\noindent \textbf{Datasets and Tasks.} We evaluate on two multi-task datasets: \textbf{PASCAL-Context}\cite{mottaghi2014role} and \textbf{Taskonomy}\cite{zamir2018taskonomy}. The PASCAL-Context includes 10,103 training images and 9,637 testing images with the five task annotation of edge detection (Edge), semantic segmentation (Seg.), human parts segmentation (H.Parts), surface normals (Norm.), and saliency detection (Sal.). The Taskonomy benchmark includes 3,793k training images and 600k testing images with 16 types of annotation. 
% \zt{Due to disk space and GPU memory limitation , we discard three types of high-dimensional annotation (points, nonfixated matches, and semantic segmentation for MSCOCO classes) that require the model to output a 3-dimensional matrix.}
% \e{Why did we not use all 16? This looks suspicious from the point of view of a reviewer, so it would be good if we could explain it.}
We use 13 annotations among them\footnote{Due to corrupt annotation for some samples, we discard three types of annotation (points, nonfixated matches, and semantic segmentation).} as our multi-task target: object classification, scene classification, depth estimation with euclidean depth, depth estimation with z-buffer depth, surface normals, curvature estimation, reshading, edge detection in 2D and 3D, keypoint detection in 2D and 3D, unsupervised segmentation in 2D and 2.5D. Details of these tasks can be found in \cite{zamir2018taskonomy}.

\noindent \textbf{Loss Functions and Evaluation Metrics.} Classification tasks and semantic segmentation use cross-entropy loss and pixel-wise cross-entropy loss respectively. Surface normals calculate the inverse of cosine similarity between the l2-normalized prediction and ground truth. Curvature estimation uses L2 loss. All other tasks use L1 loss. 

We follow previous work \cite{maninis2019attentive} to use $\Delta t_i$ to evaluate a MTL model $m$ as the average drop for task $T_i$ with respect to the baseline model $b$: $\Delta t_i = (-1)^{s_i}(M_{m,i}-M_{b, i})/M_{b, i}$ where $M_{m,i}$ and $M_{b,i}$ are the metrics of task $T_i$ for the model $m$ and $b$ respectively, and $s_i$ is $1$ if the metric is the lower the better and $0$ otherwise. We also report $\Delta t$ as the average of $\Delta t_i$ on all tasks.
For here, the baseline model $b$ is the vanilla single-task learning model. 
% Note that to better represent the performance gain, baseline model $b$ is the vanilla multi-task learning baseline model $MTL$ in taskonomy and vanilla single-task learning baseline model in PASCAL-Context. 

On the taskonomy, for depth estimation, we also report root mean square error (rmse), absolute and relative errors between the prediction and the ground truth as well as the percentage of pixels whose prediction is within the thresholds of $1.25, 1.25^2, 1.25^3$ to the ground truth following \cite{eigen2014depth}. 
We also report accuracy (Acc) for classification, L2 distance for curvature estimation, and L1 distance for all other tasks. These metrics are used to calculate $\Delta t_i$ and note that depth estimation use rmse only.

On the PASCAL-Context, we follow \cite{liang2022m} and report mean intersection over union (mIoU) for semantic and human parts segmentation, and saliency; mean error (mErr) for normals estimation, root mean square error (rmse) for depth estimation; and optimal dataset F-measure (odsF) for edge detection.

\noindent \textbf{Baselines and Competitors.} We compare with the following baselines. \textbf{STL}: vanilla single-task learning baseline that trains its own model on each task independently. \textbf{MTL}: vanilla multi-task learning baseline that all tasks share the backbone model but have separate prediction heads. For our proposed model, we also have MLP and Attn (in Table.~\ref{tab:MTL}) that represent only MoE MLP and only MoE attention networks are customized into the transformer layer respectively. \model w/ pruning (or pruning in Table.~\ref{tab:MTL}) is \model with experts removing for each specific task and we report the maximum FLOPs and Params over all tasks. 
We also compare with $M^3ViT$\cite{liang2022m} and several state-of-the-art MTL models: MTAN\cite{liu2019end}, Cross-Stitch \cite{misra2016cross} and NDDR-CNN\cite{gao2019nddr}. 
Further, we compare with \textbf{modified-MoE}: it has the same architecture as \model
% but with a modified MoE loss that apply the balanced loss\cite{zhang2022mixture} after forwarding training pairs of all tasks instead of one task. 
but without our mutual information loss. It applies the standard balanced loss~\cite{zhang2022mixture} after forward propagation of all tasks for each image instead of one task.
As a result, experts will be evenly used for all tasks instead of for every task. 

\noindent \textbf{Implementation.} 
We use ViT-base\cite{dosovitskiy2021an} and ViT-small as backbone networks on the Taskonomy and the PASCAL-Context respectively. 
We introduce MoA and MoE MLP into ViT every two layers. 
% We use MoA and MoE MLP for the layer with odd depth and even depth respectively. 
For MoA, we follow \cite{zhang2022mixture} to design the block and use 15 experts with top-k as 6 for ViT-small and 24 experts with top-k as 12 for ViT-base. For MoE MLP, we use 16 experts with top-k as 4. The task-specific heads are single linear layers on the Taskonomy and multiple layers network same as \cite{liang2022m} on the PASCAL-Context.  We set $w_{MI}=0.001$ and removed threshold $\theta=1.0\%$. 

% Our mutual information loss are summed up over all MoE modules and multiply by a constant $0.001$. The total loss is mutual information loss plus the task-specific loss for all tasks using a auto balancing weight\cite{kendall2018multi}. 

On the PASCAL-Context, the hyperparameters are the same as in $M^3ViT$\cite{liang2022m}. On the Taskonomy, we set the base learning rate to $2e-4$ with a batch size of $1,440$ and AdamW\cite{loshchilov2019decoupled} as the optimizer. The weight decay is $0.05$. We use 10 warmup epochs with 100 total training epochs and the model converges in 80 hours with 240 NVIDIA V100 GPUs.
Cosine decay\cite{loshchilov2017sgdr} is used for the learning rate schedule. 
% \model is trained with 240 NVIDIA V100 GPUs and converges in 80 hours.
%  \yk{for which dataset?}

\begin{table*}[]
    \centering
    \small
    % \hspace{-0.5in}
    \setlength{\tabcolsep}{1mm}{
    \begin{tabular}{l|c|ccccc>{\columncolor{mygray}}c|c|c} %>{\columncolor{mygray}}c
    \hline
    \multirow{2}{*}{Method} & \multirow{2}{*}{Backbone} & Seg. & Norm. & H. Parts & Sal. & Edge & $\Delta _t$ & FLOPs & Params \tabularnewline
     &  & mIoU$\uparrow$ & mErr$\downarrow$ & mIoU$\uparrow$ & mIoU$\uparrow$ & odsF$\uparrow$ & (\%)$\uparrow$ & (G)$\downarrow$ & (M)$\downarrow$  \tabularnewline
     \hline
     STL & ResNet-18 & 66.2 & 13.9 & 59.9 & 66.3 & 68.8 & 0.00 & \best{1.8} & \best{11}   \tabularnewline
    \hline
     MTL & ResNet-18 & 63.8 & 14.9 & 58.6 & 65.1 & 69.2 & $-$2.86 & \best{1.8} & \best{11} \tabularnewline
     MTAN\cite{liu2019end} & ResNet-18 & 63.7 & 14.8 & 58.9 & 65.4 & 69.6 & $-$2.39 & \best{1.8} & \best{11} \tabularnewline
    Cross-Stitch \cite{misra2016cross} & ResNet-18 & 66.1 & 13.9 & 60.6 & 66.8 & 69.9 & +0.60 & \best{1.8} & \best{11} \tabularnewline
    NDDR-CNN\cite{gao2019nddr} & ResNet-18 & 65.4 & 13.9 & 60.5 & 66.8 & 69.8 & +0.39 & \best{1.8} & \best{11} \tabularnewline
    \hline
    MTL & ViT-small & 70.7 & 15.5 & 58.7 & 64.9 & 68.8 & $-$1.77 & 4.6 & 21 \tabularnewline
    $M^3ViT$\cite{liang2022m} & MoE ViT-small & 72.8 & 14.5 & 62.1 & 66.3 & 71.7 & +2.71 & 5.2 & 42 \tabularnewline
    \hline
    % Ours-Mlp & MoE ViT-small & 73.5 & 14.6 & 62.1 & 66.4 & 71.9 & +2.86 & 5.2 & 42 \tabularnewline
    % Ours-Attn & MoE ViT-small & 73.8 & 14.5 & 62.3 & 66.6 & 71.6 & +3.13 & 5.2 & 29  \tabularnewline
    \model & MoE ViT-small & \best{74.1} & \best{13.7} & \best{62.7} & \best{66.9} & \best{72.0} & \best{+4.72} & 5.2 & 50 \tabularnewline
    \model w/ Pruning & MoE ViT-small & \best{74.1} & \best{13.7} & 62.6 & \best{66.9} & 71.9 & +4.65  & 5.2 & 22 \tabularnewline
    \hline
    % \rowcolor{mygray} Mean & - & & & & & & \tabularnewline
    
    \end{tabular}
    }
    \caption{\textbf{Quantitative Results on the PASCAL-Context.} 
    \model constantly outperform other MTL methods on all tasks.} %\gc{Give conclusion: our results are better..}
    
    \label{tab:pascal}
    % \vspace{-0.2cm}
\end{table*}

\subsection{Results on MTL}

\begin{figure}
    % \hspace{-0.5in}
  \centering
  \includegraphics[width=8.5cm]{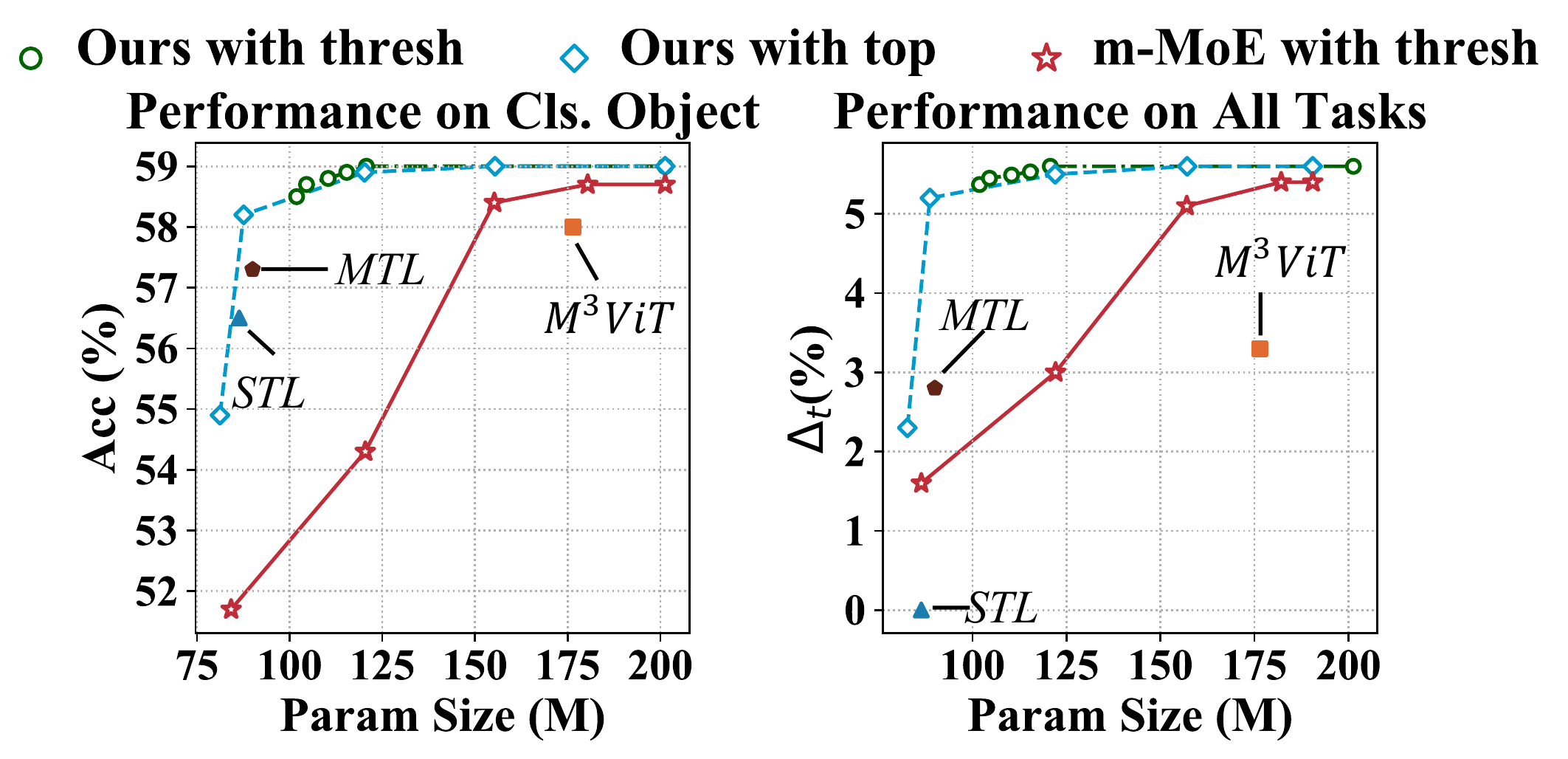}
  \caption{\textbf{Ablation study on pruning. } We explore two ways of pruning: (1) thresh then remove with $\theta$ (2) Keep the top $H\%$ experts that have the highest used frequency in each MoE module. 
  For the first way of pruning, we report results with $\theta$ as 90\%, 50\%, 20\%, 5\%, 0.1\%, and 0.0\% (no pruning). For the second way of pruning, we report results with $H\%$ as 30\%, 40\%, 60\%, 80\%, and 100\% (no pruning). We also compare our pruning with applying the same pruning strategy on modified-MoE (m-MoE).
%   \yk{can we add the performance of baseline models (like STL, MTL and M3ViT) to figures?}
%   \textcolor{red}{I think you should say something explicit about which row is considered the 'final answer'. I assume it is the detection head?}
  }
  \label{fig:prune}
%   \vspace{-0.2cm}
\end{figure}

\noindent \textbf{Efficacy.} We demonstrate the efficacy of our model in performance, computation cost, and model capacity. 
The results on the Taskonomy and the PASCAL-Context are shown in Table.~\ref{tab:MTL} and Table.~\ref{tab:pascal} respectively. Specific metrics for each task on the Taskonomy is shown in Table.~\ref{tab:metric}. 
In terms of performance, our method significantly outperforms other baselines and competitors on both datasets: we beat MTL and M$^3$ViT for over 2 points in mean $\Delta_t$ on the two datasets. On Taskonomy, we defeat MTL on all tasks, which proves the improvement is consistent. 
In terms of computation cost and model capacity, our model with ViT-Base backbone has a very low computation cost (19.7G FLOPs) while benefiting from a huge model capacity (201.3M). In comparison, MTL baselines with ViT-Base use 18.5G FLOPs with 86.4M parameters. 
Furthermore, our standalone pruned model keeps the same performance as \model for each individual task when having the same level of computation cost and model capacity as STL: 18.4 FLOPs vs. 17.7 FLOPs and 116.9M vs. 86.4M. The extra computation cost is mainly from the lightweight routing network and the extra parameters can be further removed with a higher $\theta$ as will be shown later.

\begin{figure}
    % \hspace{-0.5in}
  \centering
  \includegraphics[width=8.5cm]{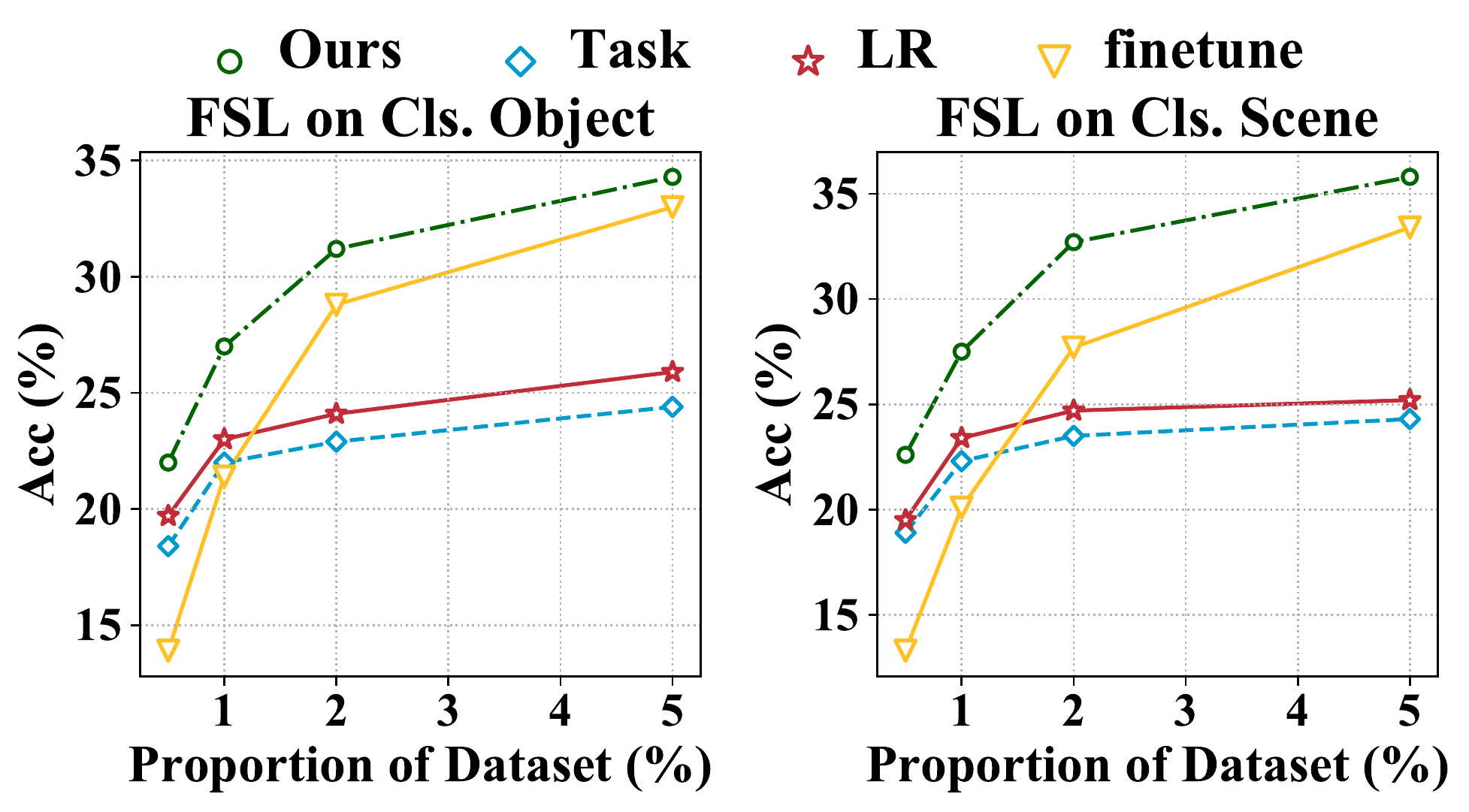}
  \caption{\textbf{Router fine-tuning can quickly learn new tasks by selecting proper experts. } We train our model on the other 11 tasks from the Taskonomy and transfer to cls. object and cls.scene with few training samples. We compare the few-shot classification accuracy with the following three baselines. (1) Fine-tuning: We fine-tune the whole model on the few training samples. (2) Task: we freeze the backbone model and only train the new task-specific head. (3) LR: the state-of-the-art few-shot learning method \cite{tian2020rethinking} based on logistic regression. We report the test accuracy when training with 0.5\%, 1\%, 2\%, and 5\% of the training set.
  }
  \label{fig:fewshot}
%   \vspace{-0.2cm}
\end{figure}

% \textbf{Flexibility.} We demonstrate the flexibility from two aspects: (1) our model is capable of removing experts and adapt extract a compact standalone model for individual task based on the trade-offs between parameters size and performance; (2) the model can be quickly adapted to a new task or scenario with a few examples by only updating the routing network.

\noindent \textbf{Ablation study on MoE Mlp and MoE Attention.}
As shown in Table.~\ref{tab:MTL}, we report results (MLP and Attn in Table.~\ref{tab:MTL}) where we only introduce MoE into MLP and attention networks. Both ways of adding experts can improve $>1.0\%$ in $\Delta_t$ compared to MTL. By combining them, \model gets the best result and further boost 2 points in $\Delta_t$. This demonstrates that introducing MoE and increasing model capacity in both attention and MLP network can increase the performance. 

\subsection{Experts, Tasks, and Pruning}

\label{relation}
\begin{figure}
    % \hspace{-0.5in}
  \centering
  \includegraphics[width=8.5cm]{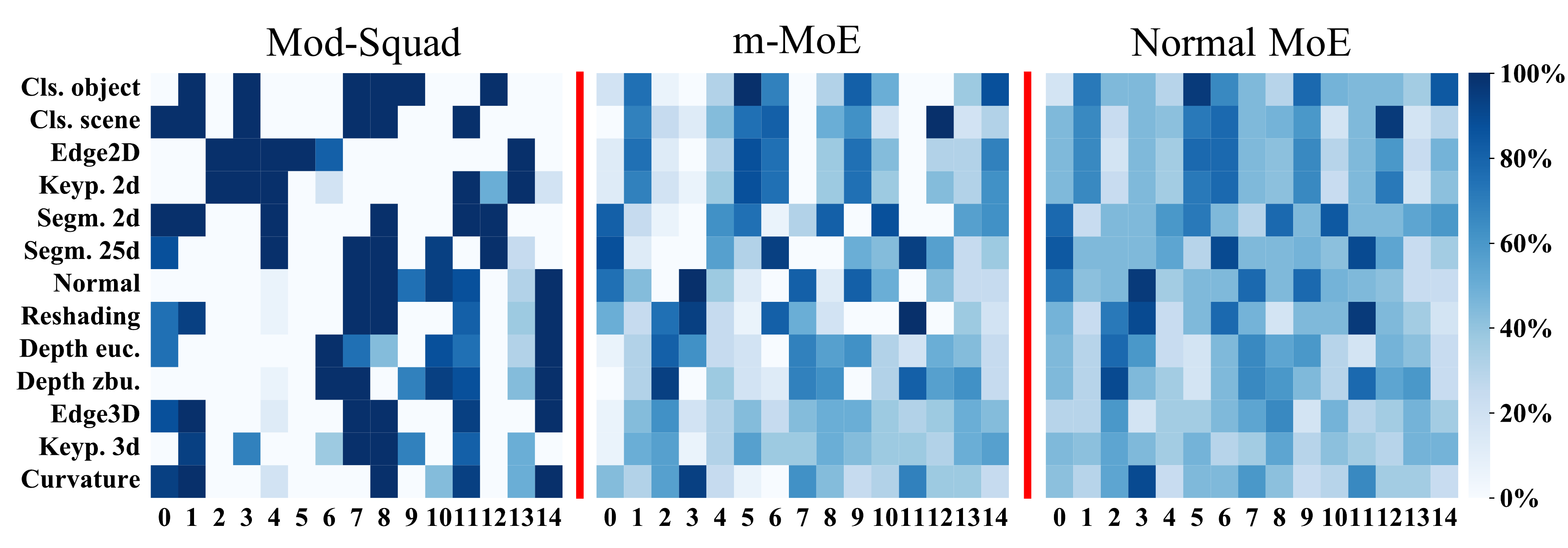}
  \caption{\textbf{Visualization of the frequency that experts being selected for each task. } We visualize the activation frequency of a MoE attention module in the 6-th transformer block with 15 experts and top-k as 6. The y-axis represents the tasks and the x-axis represents the 15 experts. We compare the visualization of \model to m-MoE and normal MoE. All three methods have the exact same MoE module but with different MoE losses. Our frequency map is much \textbf{sharp} and \textbf{sparse} than other methods.
%   \textcolor{red}{I think you should say something explicit about which row is considered the 'final answer'. I assume it is the detection head?}
  }
  \label{fig:expert_task}
%   \vspace{-0.2cm}
\end{figure}

\noindent \textbf{Relation between experts and tasks.} As shown in Fig.~\ref{fig:expert_task}, we visualize the frequency of experts being selected for each task. The x-axis and y-axis represent experts and tasks respectively. Experiments are conducted on the Taskonomy with all 13 tasks using MoE ViT-Small as the backbone. The visualization is for the MoE attention module in the 6th transformer block. We also compare with modified-MoE and Normal MoE which have different MoE losses but the exact model architecture. From the figure, we observe that our expert activation map is sharper and more sparse than the two comparisons, which aligns with our key motivation: a sparse but strong dependence between experts and tasks helps MTL.
% Based on the sparse depence, we can further remove experts and extract sub-network for individual task.

\noindent \textbf{Extracting sub-network for an individual task. } 
As introduced in \cref{sec:pruning}, we extract a small sub-network from \model for an individual task. Specifically, we explore two ways of removing experts as follows. (1) Thresh and remove: we simply remove all experts that have an usage frequency lower than $\theta$ for the specific task. Note that some MoE modules could have fewer than Top-K experts after removing if most of the experts have a low usage frequency. In that case, we reduce the top-k of that MoE module to the number of experts it keeps. (2) Keep the top: we keep the top $H\%$ experts in each MoE module that have the highest usage frequency. 

The results are shown in Fig. \ref{fig:prune}. For the first way of removing experts, we try $\theta$ as 90\%, 50\%, 20\%, 5\%, 0.1\%, and 0\% (no removing). For the second way, we try $H\%$ as 50\%, 20\%, 5\%, and 0\% (no removing). For both removing strategies, we compare with STL, MTL, and M$^3$ViT. From the figure, we notice several interesting observations: (1) \model can remove the majority of extra experts than a normal ViT-Base (116.9M vs. 90.0M in model parameters) with a tiny performance lost ($<0.3\%$ in $\delta_t$) and still better than competitors. (2) Only keeping the top 40\% of experts still give us the same performance (5.5\% in $\delta_t$ while the best is 5.6\%). (3) The performance of modified-MoE significantly drops when removing more experts, which prove the effectiveness of our mutual information loss.

% In a short word, our model is sparse in image-level and task-level but dense in dataset-level, which allow extracting  a seriels of standalone small sub-network for individual task once training.

\noindent \textbf{Fine-tuning the router network. }
Another interesting property of \model is that we can quickly adapt to new tasks by only tuning the lightweight routing network and the task-specific head with all other parameters frozen. We refer to this technique as router fine-tuning.
Router fine-tuning can be generalized to any MoE network when they need lightweight tuning with limited budgets in dataset size, computation cost, or training time. 
% that can applied to quickly fine-tuning for new tasks and scenarios with limited budgets in training samples, computation cost, training time, and etc. 

As shown in Fig. \ref{fig:fewshot}, we explore router fine-tuning. We first pre-train our model on 11 tasks on the Taskonomy except for cls. object and cls. scene as the target of new tasks. We compare different ways of fine-tuning with limited training examples. We report performance using 0.5\%, 1\%, 2\%, and 5\% of the dataset to learn the new tasks. The router fine-tuning strategy is compared with several baselines as follows. (1) Fine-tuning: fine-tune the whole model and learn the new task-specific head. (2) Task: freeze the backbone model and only learn the new task heads. (3) We follow the state-of-the-art few-shot learning method \cite{tian2020rethinking} based on logistic regression to fine-tune the model.

From the figure, we find that the router fine-tuning strategy surpasses other baselines constantly on both tasks with different proportions of the training set. These results show that \model can be quickly adapted for various purposes with router fine-tuning.

\begin{figure}
    % \hspace{-0.5in}
  \centering
  \includegraphics[width=8.5cm]{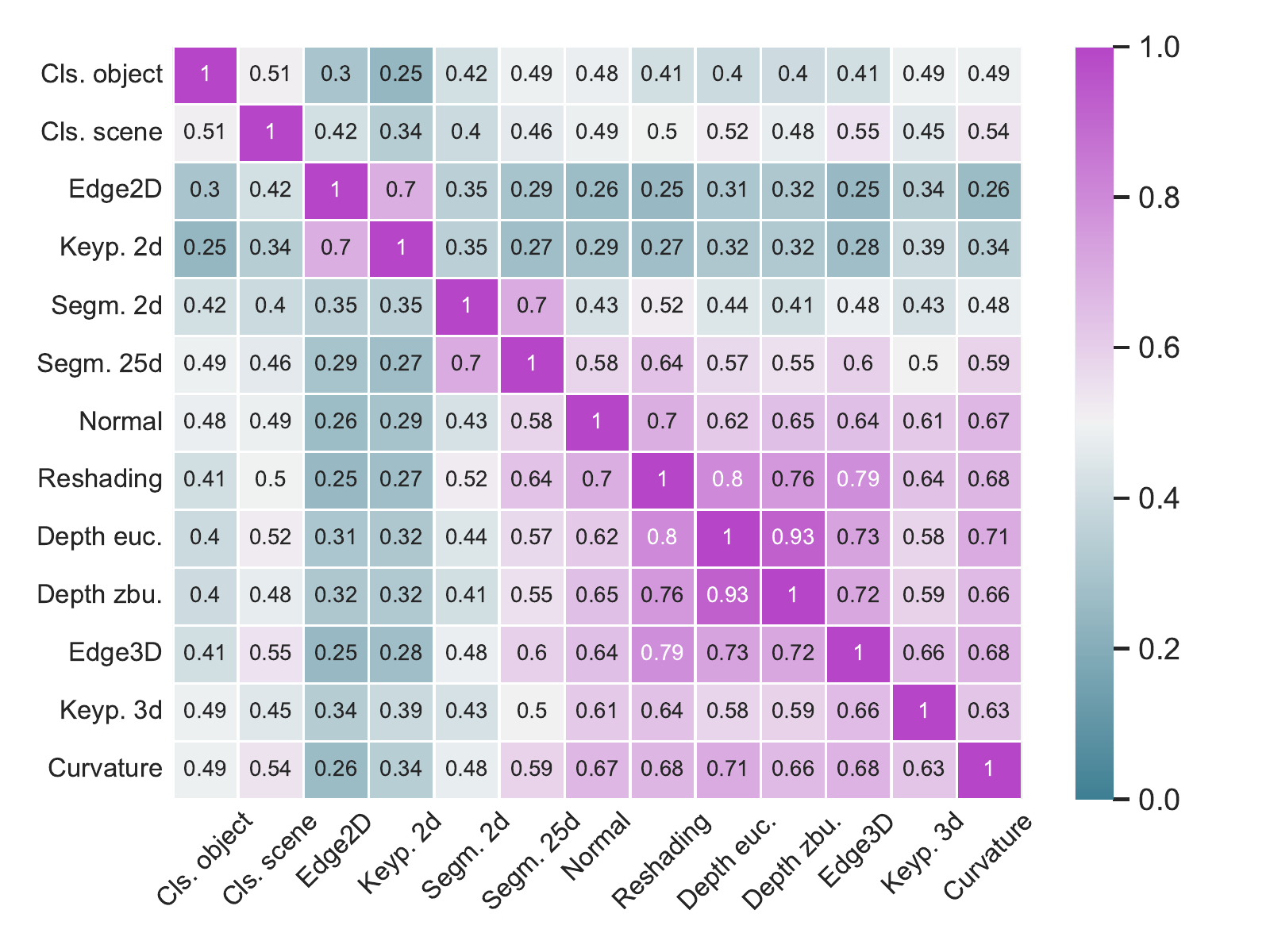}
  \caption{\textbf{Task relation from \model.} We evaluate the similarity between tasks as the mean of the percentage of experts that they are sharing with the same input. 
  }
  \label{fig:task}
  \vspace{-0.05in}
\end{figure}

\noindent \textbf{Task Relation. } \model can not only model the task relation implicitly like other multi-task models but also visualize it explicitly.
% is also capable of learning and utilizing the relation between tasks. Unlike any multi-task model that model the relation implicitly, our model can explicitly define and visualize the relation between tasks. 
We define the similarity between tasks as the mean of the percentage of experts that they are sharing given the same input. If two tasks are sharing more experts than other pairs of tasks, they are considered to be more related. 
This definition may not be perfectly accurate but is based on one simple rule: related tasks are more likely to share experts than unrelated tasks. 
% With this definition, task relation can be visualized by our model.

As shown in Fig.~\ref{fig:task}, \model visualizes task relations in a correlation matrix with our new definition of task similarity. We notice that some of the structures among tasks are interesting: the 3D tasks including Normal, Reshading, two depth tasks, Edge3D, Keyp. 3D and curvature are grouped together; closed relation exists among two segmentations tasks and among two two depth tasks; Edge2D and Edge3D are not closed in the visualization. It demonstrates \model can also be used as a visualization tool to explore the structure among tasks.
% The visualization tool can be used to explore more structure among various tasks.
% We believe this visualization will help us better understand multi-task learning and inspire us for better solutions. 
% We define the similarity between two task as $\frac{a}{b}$. Here $a$ and $b$ is the number of experts that are activated for both tasks and the number of experts that are activated at least for one of the two tasks when giving the same input image respectively.  The motivation is that more similar tasks prefer more on using the same experts. 

\section{Conclusion}
In this work, we propose \model, a modular multi-task learner based on mixture-of-experts and a novel loss to address the gradient conflicts among tasks. We demonstrate its potential to scale up in both model capacity and target task numbers while keeping the computation cost low.
It is noteworthy that \model can be scaled down in model size with no performance drop for specific purposes.
Future work could extend \model to a large variety of tasks and scenes not only in the vision domain but also in other modalities (e.g., text and audio). We hope \model will become an important building block of future efficient and  modular foundation models.

% \gc{also talk about limitation and future works}

% We remove the experts if its activate frequency is less than a threshold $\theta$ and note that if a layer after pruning have less than $k$ experts, we keep the $k$ most frequent used experts. Here $k$ is the number of selected experts in a noisy top-k routing network. We also try a smaller $k$ for the extracted sub-network. 

%%%%%%%%% REFERENCES
{\small
\bibliographystyle{ieee_fullname}
% \bibliography{egbib}
\bibliography{cvprbib}
}

\newpage
\appendix
\onecolumn

\setcounter{table}{0}
\setcounter{figure}{0}
\renewcommand{\thetable}{A\arabic{table}}
\renewcommand{\thefigure}{A\arabic{figure}}
\renewcommand{\thesubsection}{A\arabic{subsection}}

\section*{Appendix}

\renewcommand\twocolumn[1][]{#1}%
\maketitle
\begin{center}
    \vspace{-0.1in}
    \captionof{table}{
    Metric for more tasks on the taskonomy dataset. The experiment section in the paper demonstrates what metric we use for each task.
    }
    \centering
    {
    % \footnotesize 
    %\scalebox{1}{ 
    %\tabcolsep=0.08cm
    \begin{tabular}{c|c|cc|ccc|c|c|c}
\toprule
\multirow{3}{*}{Model} & \multicolumn{6}{c|}{Depth Zbu.}     & \multicolumn{1}{c|}{Edge2D} &   \multicolumn{1}{c|}{Keyp.3D} & \multicolumn{1}{c}{Segm. 2.5D} \\ \cline{2-10}
                    & \multirow{2}{*}{$RMSE \downarrow$} & \multicolumn{2}{c|}{Error $\downarrow$} & \multicolumn{3}{c|}{$\delta$, within $\uparrow$} &    \multirow{2}{*}{L1 dis. $\downarrow$} &  \multirow{2}{*}{L1 dis. $\downarrow$} &    \multirow{2}{*}{L1 dis. $\downarrow$}           \\ \cline{3-7}
                                    &          &        Abs.  &     Rel.   &   1.25 & $1.25^2$ & $1.25^3$   &    & &                               \\
\midrule                     

STL & 6.90 & 0.086 & 1.71 & 92.6 & 97.0 & 98.7 & 0.00500 & 0.072 & 0.155 \\
MTL & 6.75 & 0.083 & 1.30 & 93.1 & 96.8 & 98.9 & 0.00497 & 0.071 & 0.152 \\
$M^3ViT$  & 6.73 & 0.081 & 1.30 & \best{93.4} & 97.3 & 98.9 & 0.00490 & 0.070 & 0.150 \\
\hline
\model & \best{6.70} & \best{0.080} & \best{1.28} & 93.3 & \best{97.5} & \best{99.1} & \best{0.00482}     & \best{0.068} & \best{0.147}       \\
% Ours-Pruning &  & & & & & & & & & & & & &  \\

\bottomrule
\end{tabular}
    }
    %}
    \label{tab:metric}

% \end{table*}
\end{center}
%}]

\begin{figure}[!hbp]
    % \hspace{-0.5in}
  \centering
  \caption{\textbf{Visualization of the pruning results on each task. } We set $\theta=0.1\%$ and do the pruning. Every tasks keep the same performance while reducing the parameters size from over 200M to lower than 120M.
  }
  \vspace{-0.1in}
  \includegraphics[width=13cm]{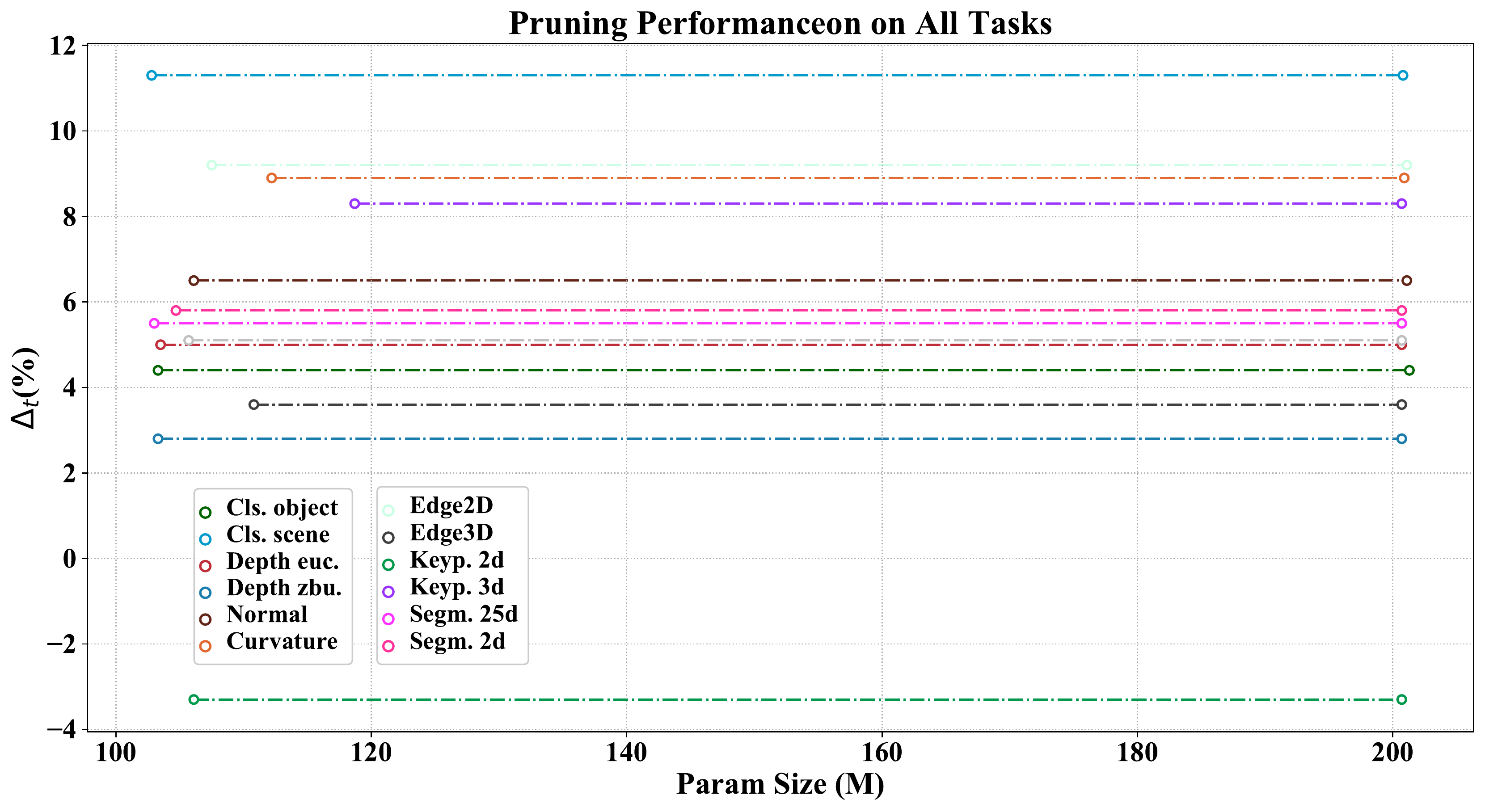}
  
  \label{fig:all_layer}
   \vspace{-0.1in}
\end{figure}

\subsection{Metric for more task on the Taskonomy. } We provide more results with the task metric in Table.~\ref{tab:metric}. Other tasks are shown in the paper already. \model consistently have the best performance on these metrics.

\subsection{Pruning on different tasks.}
We show the pruning results for all tasks on the Taskonomy in Figure.~\ref{fig:all_layer}. We set $\theta=0.1\%$ and removing all experts have an activation frequency lower than $\theta$. Every tasks keep the same performance while reducing the parameters size from over 200M to lower than 120M. This demonstrate the effectiveness of \model on all tasks.

\subsection{Ablation on Top-$K$.}
As showin in Table.~\ref{tab:topk}, we explore the effect on Top-$K$ for both MoE attention layer and MoE MLP layer. The experiment setting is the same as in \cref{sec:expert}. To control the FLOPs to be the same for different Top-$K$, the hidden dimension of attention experts and the mlp experts is divided by $K$. All experiments have the same parameter size and the same FLOPs. 

\subsection{Ablation on number of experts.}
\label{sec:expert}
As showin in Table.~\ref{tab:expert_number}, we explore the effect on number of experts $E$ for both MoE attention layer and MoE MLP layer. For quick experiments, we use ViT-small as the backbone network. In default, we add MoE attention layer with 15 experts and Top-$K$ as 6 as well as MoE MLP layer with 16 experts and toop-k as 4. The MoE modules are added at every layer. We use $\Delta_t$ to evaluate the multi-task performance and report the parameters size. The $\Delta_t$ compare various version of models to the vanilla single task learning baseline with ViT-small.
The experiments show that increasing experts number bring extra performance but the gain gradually diminish as the $E$ goes up. This conclusion hold true for both MoE attention layer and MoE MLP layer.

\subsection{More visualization on \model.}
\begin{figure}[!hbp]
    % \hspace{-0.5in}
  \centering
  \caption{\textbf{Visualization of the frequency that experts being selected for each task in the MoE MLP layer. } We visualize Mod-Squad based on ViT-small. The activation frequencies of MoE MLP modules are shown in all
 transformer blocks with 16 experts and Top-$K$ as 4.
 The y-axis represents the tasks and the x-axis represents the experts. It demonstrates the sparsity of \model in all layers for all tasks.
  }
  \includegraphics[width=\linewidth]{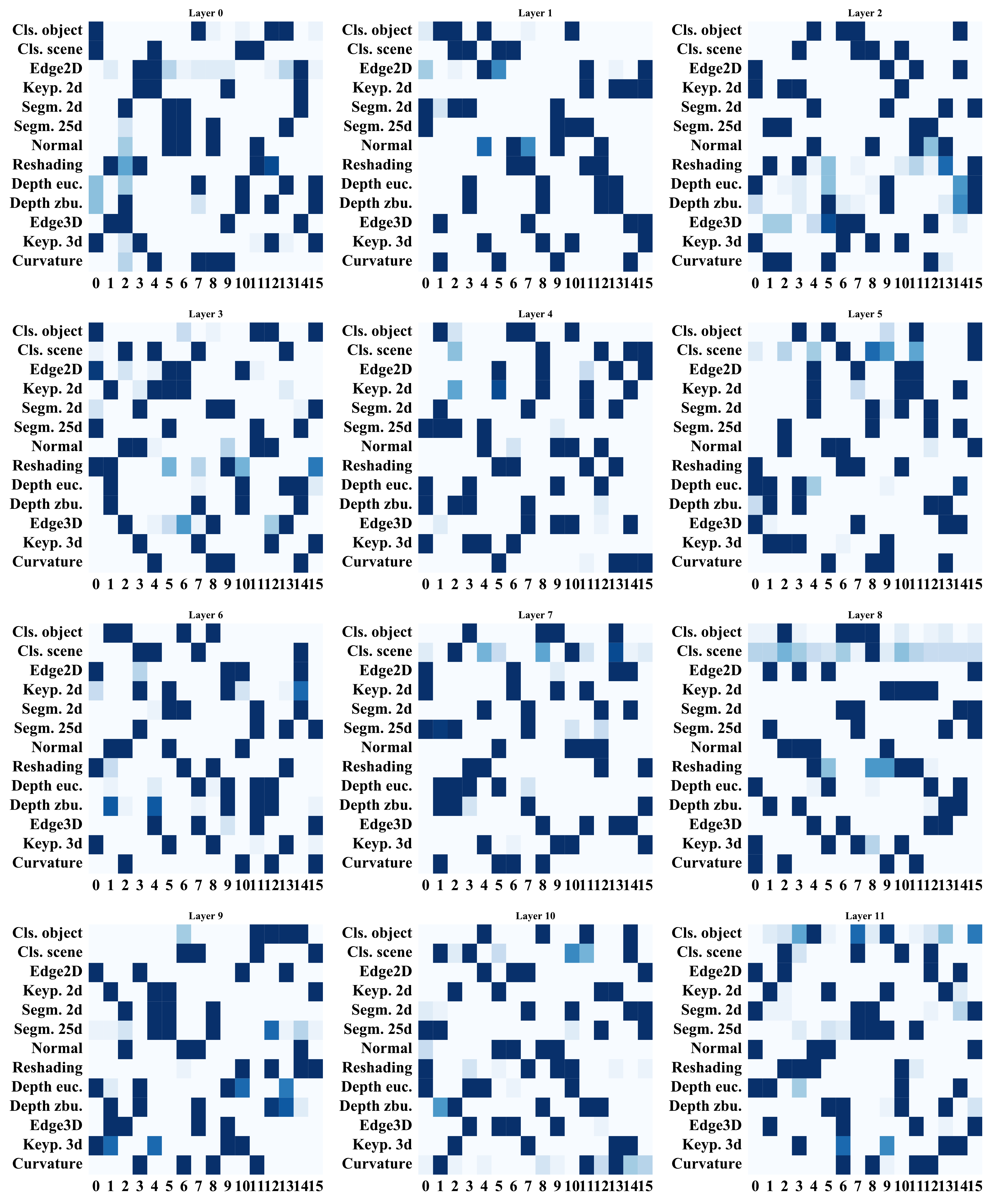}
  
  \label{fig:mlp}
%   \vspace{-0.2cm}
\end{figure}

We visualize \model based on ViT-small as shown in Figure.~\ref{fig:all_layer} and Figure.~\ref{fig:mlp}. To better understand the relation between experts and task in all layers, we insert MoE attention layer and MoE MLP layer on every transformer block. 
In Figure.~\ref{fig:all_layer}, the activation frequencies of MoE attention modules are shown in all transformer blocks with 15 experts and Top-$K$ as 6. In Figure.~\ref{fig:mlp}, the activation frequencies of MoE MLP modules are shown in all transformer blocks with 16 experts and Top-$K$ as 4.
Both visualization demonstrates the sparsity of \model in all layers for all tasks.

\subsection{Task relation from different layers of \model.}
\begin{figure}[!hbp]
    % \hspace{-0.5in}
  \centering
  \caption{\textbf{Visualization of task similarity from the first, the middle, and the last layers. } For each layer $L_i$, we evaluate the similarity between a task pair as the mean of the percentage of experts in $L_i$ that task pairs are sharing with the same input. 
  }
  \includegraphics[width=\linewidth]{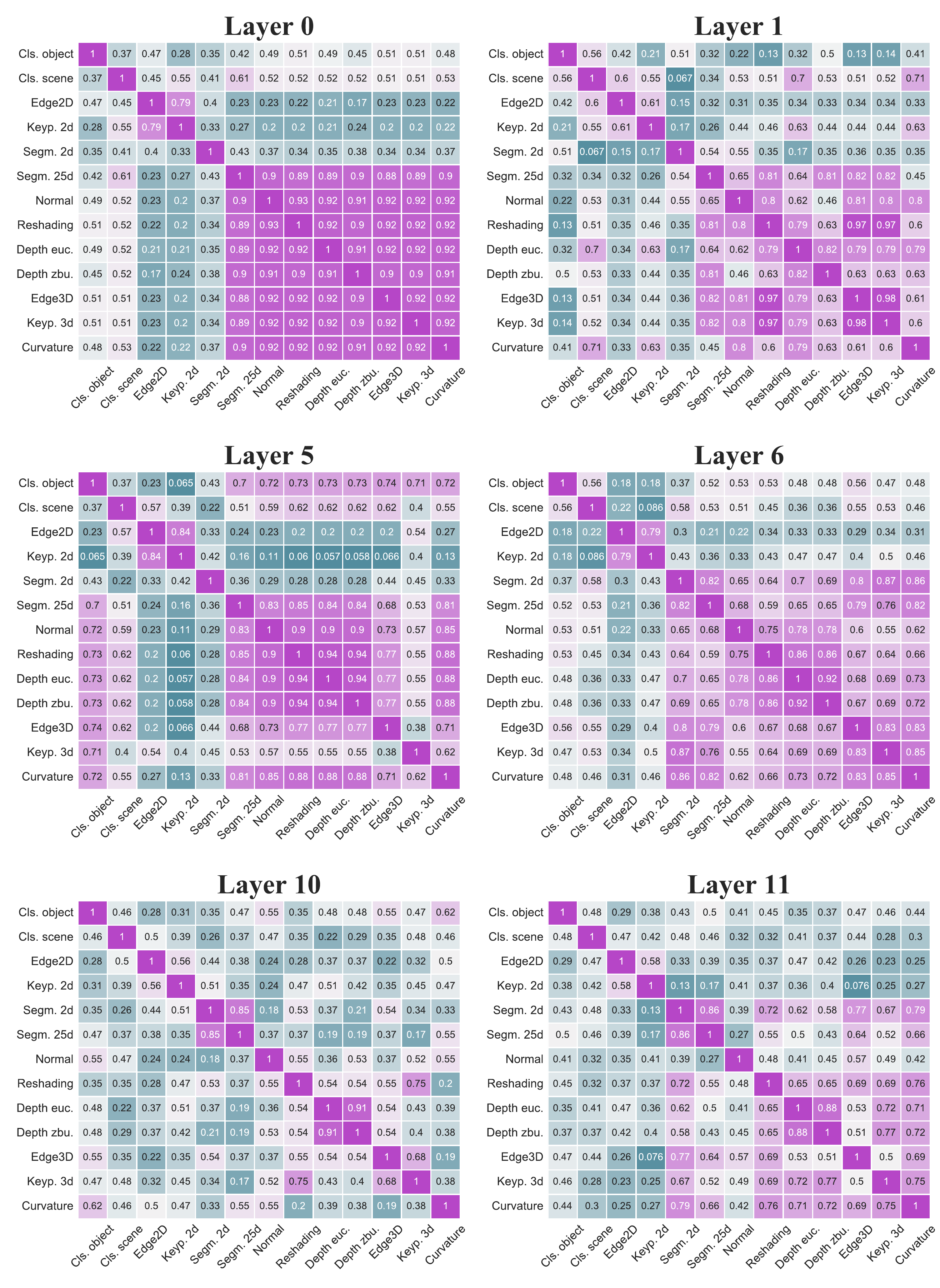}
  
  \label{fig:layer_relation}
%   \vspace{-0.2cm}
\end{figure}

In the paper, we define the similarity between tasks as the mean of the percentage of experts that they are sharing given the same input. We put all experts into the calculation of task similarity. However, we can also calculate the task similarity for each layer, by only put the experts from that specific layer into the calculation. As shown in Figure.~\ref{fig:layer_relation}, we visualize the task similarity for the first two layers, the middle two layers, and the last two layers of \model. \model is trained on the Taskonomy with ViT-base as the backbone and follow the same setting in our Taskonomy experiments in the paper. We notice that (1) all 3d tasks are very similar in the first layer; (2) all tasks are more similar to classification tasks in the middle two layers; (3) there is less similarity between tasks in the last two layers. 

\begin{table*}[t]
    \centering
    \small
    \caption{\textbf{Ablation study of Top-$K$ on MoE attention layer and MoE MLP layer.}}
    \label{tab:topk}
    \setlength{\tabcolsep}{0.9mm}{
    \begin{subtable}[t]{0.47\linewidth}
        \centering
        \small    
        \begin{tabular}[h]{c|c|c|c|c}
        
          &    FLOPs(G)     & Params(M) & Hidden Dim &  $\Delta_t$  \\ \hline
        {K=2}        &   5.2   &    75 & 192 &  -0.4          \\
        {K=4}        &   5.2   &    75 & 96 &     1.3          \\
        \rowcolor{mygray}
        {K=6}        &   5.2   &    75 &  64 &   \textbf{3.5}              \\
        {K=9}        &  5.2    &    75 &  42 &   3.1             \\ 
        {K=12}       &  5.2    &    75 &  32 & 2.2    \\
        \end{tabular}
        \caption{\textbf{Ablation on Top-$K$ in the MoE attention layer.}}
        \label{tab:aa}
    \end{subtable}%
    \hspace*{1.5em}
   \begin{subtable}[t]{0.47\linewidth}
        \centering
        \small    
        \begin{tabular}[h]{c|c|c|c|c}
        
          &    FLOPs(G)     & Params(M) & Hidden Dim & $\Delta_t$   \\ \hline
        {K=1}        &   5.2   &    75 &  1536 & 3.3          \\
        {K=2}        &   5.2   &    75 &  768  &    3.3          \\
        \rowcolor{mygray}
        {K=4}        &   5.2   &    75 &  384  &   \textbf{3.5}              \\
        {K=6}        &  5.2    &    75 &  256  &   3.2             \\ 
        {K=8}       &  5.2    &    75  &  192  & 3.1    \\
        \end{tabular}
        \caption{\textbf{Ablation on Top-$K$ in the MoE MLP layer.}}
        \label{tab:bb}
    \end{subtable}%
    }

\end{table*}

\begin{table*}[t]
    \centering
    \small
    \caption{\textbf{Ablation study of expert number E on MoE attention layer and MoE MLP layer.}}
    \label{tab:expert_number}
    \setlength{\tabcolsep}{0.9mm}{
    \begin{subtable}[t]{0.47\linewidth}
        \centering
        \small    
        \begin{tabular}[h]{c|c|c}
        
          &   Params(M) & $\Delta_t$  \\ \hline
        {E=6}        &   66   &     2.0           \\
        {E=9}        &   69   &     2.7              \\
        {E=12}       &  72    &     3.2              \\ 
        \rowcolor{mygray}
        {E=15}       &  75  &  \textbf{3.5}     \\
        {E=18}       &  78  & \textbf{3.5}     \\
        \end{tabular}
        \caption{\textbf{Ablation on $E$ in the MoE attention layer.}}
        \label{tab:a}
    \end{subtable}%
    \hspace*{1.5em}
    \begin{subtable}[t]{0.47\linewidth}
        \centering
        \small
        \begin{tabular}[h]{c|c|c}
        
          &   Params(M) & $\Delta_t$  \\ \hline
        {E=4}        &   33   &     2.2           \\
        {E=8}        &   47   &     2.7              \\
        {E=12}       &  61    &     3.3              \\ 
        \rowcolor{mygray}
        {E=16}       &  75  &  \textbf{3.5}     \\
        {E=20}       &  90  & \textbf{3.5}     \\
        \end{tabular}
        \caption{\textbf{Ablation on E in the MoE MLP layer.}}
        \label{tab:b}
    \end{subtable}%
    }

\end{table*}

\begin{figure}[!hbp]
    % \hspace{-0.5in}
  \centering
  \caption{\textbf{Visualization of the frequency that experts being selected for each task in the MoE attention layer. } We visualize Mod-Squad based on ViT-small. The activation frequencies of MoE attention modules are shown in all
 transformer blocks with 15 experts and Top-$K$ as 6.
 The y-axis represents the tasks and the x-axis represents the experts. It demonstrates the sparsity of \model in all layers for all tasks.
  }
  \includegraphics[width=\linewidth]{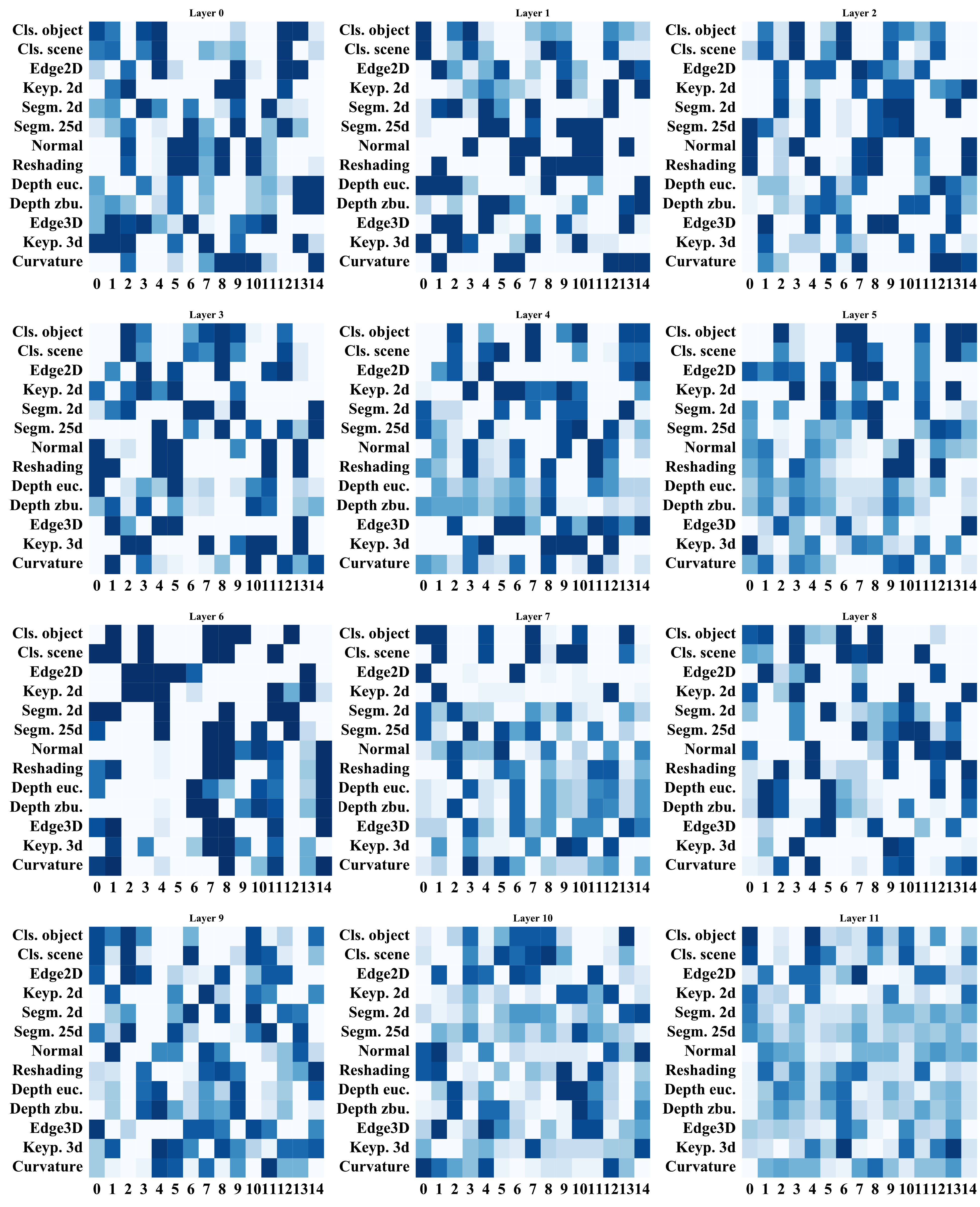}
  
  \label{fig:all_layer}
%   \vspace{-0.2cm}
\end{figure}

\end{document}